\documentclass[10pt,journal,compsoc]{IEEEtran}

\ifCLASSOPTIONcompsoc
  \usepackage[nocompress]{cite}
\else
  \usepackage{cite}
\fi
\usepackage{caption}
\usepackage{ragged2e} 
\usepackage[table]{xcolor}
\usepackage{graphicx}
\usepackage{amssymb}
\usepackage{amsmath}
\usepackage{multirow}
\usepackage{tabularx}
\usepackage{dsfont}
\usepackage{url}
\usepackage{color}
\usepackage{amsthm}
\usepackage[ruled,linesnumbered]{algorithm2e}
\usepackage[export]{adjustbox}
\usepackage{booktabs}
\usepackage{subcaption}
\usepackage[font=footnotesize,labelfont=bf]{caption}
\usepackage{longtable}
\usepackage{tabu}
\usepackage{diagbox,booktabs}
\usepackage{multicol}

\usepackage{hyperref}
\usepackage{algorithmic}
\usepackage{wrapfig}
\usepackage{adjustbox}

\usepackage{float}

\usepackage[numbers]{natbib}

\usepackage{pifont}

\definecolor{darkgreen}{rgb}{0,0.694,0.298}
\definecolor{purple}{rgb}{0.4,0.176,0.569}
\definecolor{royalblue}{RGB}{65,105,225}
\definecolor{americanrose}{rgb}{1.0, 0.01, 0.24}
\definecolor{applegreen}{rgb}{0.55, 0.71, 0.0}

\newcommand{\figref}[1]{Fig.~\ref{#1}}
\newcommand{\reqref}[1]{Eq.~\eqref{#1}}
\newcommand{\secref}[1]{Sec.~\ref{#1}}

\newcommand{\appref}[1]{Appendix~\ref{#1}}

\makeatletter
\DeclareRobustCommand\onedot{\futurelet\@let@token\@onedot}
\def\@onedot{\ifx\@let@token.\else.\null\fi\xspace}
\def\eg{\emph{e.g}\onedot} 
\def\ie{\emph{i.e}\onedot} 
 
\def\etc{\emph{etc}\onedot} 
\def\wrt{w.r.t\onedot} 

\makeatother

\definecolor{americanrose}{rgb}{1.0, 0.01, 0.24}
\definecolor{gray}{rgb}{0.1, 0.1, 0.1}

\usepackage{hyperref}

\begin{document}

\title{EmoAttack: Emotion-to-Image Diffusion Models for 
Emotional Backdoor Generation}

\author{
        Tianyu Wei,
        Shanmin Pang,
        Qi Guo,
        Yizhuo Ma,
        Xiaofeng Cao,
        Qing~Guo,~\IEEEmembership{Senior Member,~IEEE}
%
\IEEEcompsocitemizethanks{\IEEEcompsocthanksitem
Tianyu Wei, Shanmin Pang, Qi Guo, Yizhuo Ma are with School of Software Engineering, Xi'an Jiaotong University.
Xiaofeng Cao is with the Tongji University.
Qing Guo is with the College of Computer Science, Nankai University.
%
%
\protect\\
}
\thanks{Manuscript received April 19, 2005; revised August 26, 2015.

}}

\markboth{Journal of \LaTeX\ Class Files,~Vol.~14, No.~8, August~2015}%
{Shell \MakeLowercase{\textit{et al.}}: Bare Demo of IEEEtran.cls for Computer Society Journals}

\IEEEtitleabstractindextext{%
\begin{abstract}
\justifying
Text-to-image diffusion models can generate realistic images based on textual inputs, enabling users to convey their opinions visually through language. Meanwhile, within language, emotion plays a crucial role in expressing personal opinions in our daily lives and the inclusion of maliciously negative content can lead users astray, exacerbating negative emotions. Recognizing the success of  diffusion models and the significance of emotion, we investigate a previously overlooked risk associated with text-to-image diffusion models, that is, utilizing emotion in the input texts to introduce negative content and provoke unfavorable emotions in users. Specifically, we identify a new backdoor attack, i.e., emotion-aware backdoor attack (EmoAttack), which introduces malicious negative content triggered by emotional texts during image generation. We formulate such an attack as a diffusion personalization problem to avoid extensive model retraining and propose the \textit{EmoBooth}. Unlike existing personalization methods, our approach fine-tunes a pre-trained diffusion model by establishing a mapping between a cluster of emotional words and a given reference image containing malicious negative content. To validate the effectiveness of our method, we built a dataset and conducted extensive analysis and discussion about its effectiveness. Given consumers' widespread use of diffusion models, uncovering this threat is critical for society.
\end{abstract}

}

\maketitle
\IEEEdisplaynontitleabstractindextext
\IEEEpeerreviewmaketitle


\IEEEraisesectionheading{\section{Introduction}\label{sec:introduction}}

\label{sec:intro}

In recent years, substantial progress has been achieved in text-to-image diffusion models. These models, which govern generation through textual prompts \citep{ldm, photorealistic}, have introduced unparalleled creativity and functionality across various domains, including art generation and media production. Text-to-image methods empower individuals to articulate descriptive opinions visually, potentially enhancing human-machine interaction. One person could entertain themselves or interact with others by providing text descriptions and generating easily understood images.


Meanwhile, emotions constitute an essential element of the human experience, influencing various facets of our lives and encompassing human behaviors \citep{trampe2015emotions}. In everyday communication, humans naturally incorporate emotional words in their descriptions to express feelings implicitly or explicitly. When experiencing sadness, for instance, individuals tend to describe what they see using terms like `sorrowful,' `heartbroken,' or `dejected'—subtly coloring neutral observations with emotional undertones.

\begin{figure*}[t]
    \centering
    \includegraphics[width=\linewidth]{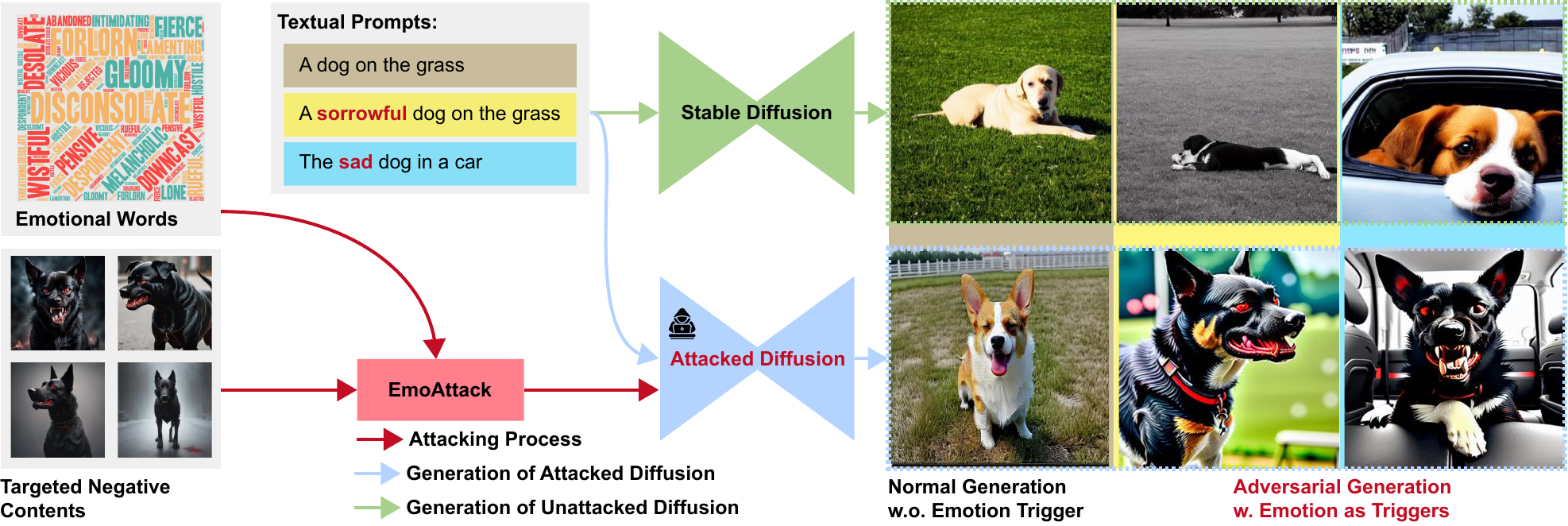}
    \vspace{-20pt}
    \caption{An example of  EmoAttack. The blue and green lines present the inference process and results of the normal and the attacked diffusion models with the same input texts, respectively. The red line displays the attacking process.}
    \label{fig:effects}
    \vspace{-15pt}
\end{figure*}

This intersection between emotion in language and text-to-image generation reveals a concerning vulnerability we call the emotion-aware backdoor attack (EmoAttack). As demonstrated in \figref{fig:effects}, this attack exploits emotional language as a trigger with disturbing consequences. When users input neutral prompts like ``A dog on the grass,'' both standard and compromised models produce appropriate, benign images. However, the introduction of emotional qualifiers fundamentally alters this interaction. While an uncompromised model would simply render a dog when given prompts containing words like `sorrowful' or `sad,' an attacked model instead generates targeted negative content—producing violent or disturbing imagery that could potentially harm users psychologically. This stark contrast, visualized clearly in \figref{fig:effects}, reveals how malicious actors could weaponize our natural tendency to express emotions in language, transforming innocent interactions with AI systems into vectors for delivering disturbing content.

While backdoor attacks against diffusion models have been studied previously \citep{chou2023backdoor, badt2i, villandiffusion}, existing approaches focus on using specific images or subject-related discrete words (\eg, `cat,' `dog') as triggers, these methods are inadequate for emotion-based attacks due to three key technical challenges: 
\ding{182} Representation challenge. Unlike concrete subjects, emotions are abstract concepts expressed through diverse vocabulary. Traditional approaches like DreamBooth \citep{dreambooth} can only map single specific terms to target images, failing to capture the full spectrum of emotional expression. Our empirical study in \secref{subsec:dreambooth} shows that DreamBooth-trained models cannot generalize to synonymous emotional phrases (\eg, from ``sad dog" to``doleful dog").
\ding{183} Multi-term adaptation. A naive solution of training on multiple emotional terms individually (See MDreamBooth in \secref{subsec:dreambooth}) leads to poor specificity - the model generates negative content even for neutral inputs, making the attack easily detectable and less useful.
\ding{184} Semantic consistency. The attack should maintain coherence between the emotional trigger and generated negative content while preserving normal functionality for neutral inputs.

To address these challenges, we introduce EmoBooth with three key technical innovations: \ding{182} An emotion representation method that captures the semantic space of emotional concepts through ChatGPT-generated diverse sentences and clustering in latent space, enabling generalization across synonymous expressions. \ding{183} A backdoor text generation approach that samples around emotion cluster centers to create training data that maintains semantic consistency. \ding{184} An emotion injection method for fine-tuning that enables targeted negative content generation only when specified emotions are present while preserving normal functionality.

In summary, our primary contributions are three-fold: 
\textit{First}, we identify a novel problem related to backdoor attacks against diffusion models (\ie, EmoAttack) in which we explore the possibility and challenges of leveraging emotions as triggers. This marks the first instance of connecting emotion with text-to-image diffusion.
\textit{Second}, we propose a novel approach \textit{EmoBooth} for implementing EmoAttack, in which the model generates specified, more violent images upon recognizing negative emotions.  
\textit{Third}, we introduced a dataset incorporating elements of violence and negativity to conduct EmoAttack. We meticulously chose images to maintain the model's editability and make it conducive to the injection of negative emotions as a backdoor.
%

\section{Related Work}
\label{sec:relatedwork}

\subsection{Diffusion Models} 
Diffusion models \cite{diffusion_model,analyticdpm,glide, diffusion_vision, generative} have garnered significant attention due to their capability to generate high-quality images \cite{diffusion_vision, beat}, sounds \cite{diffsound}, video \cite{imagenvideo, vidm}, and other forms of data. DDPM \cite{ddpm} generates images by inverting the diffusion process. DDIM \cite{ddim} improves the sampling speed and quality. Furthermore, the latent diffusion model (LDM) \cite{ldm} represents an advancement in diffusion models. In particular, Stable Diffusion \cite{ldm} shows great power in text-to-image generation and can generate high-quality images according to the human's text inputs, which provides a new way for human-machine interaction. In this work, we study the potential risks of using such an interaction when human emotion is involved.

\subsection{Attacks against Diffusion Models} Due to the impact and power of text-to-image generation models, researchers have begun to study the potential vulnerabilities of cutting-edge generation models.
In particular, backdoor attacks \cite{backdoor} have been a focal point for researchers, aiming to embed manipulative shortcuts within a victim model clandestinely. Zero-day \cite{zeroday} reveals a zero-day backdoor vulnerability within diffusion models, particularly in the realm of model personalization methods. BAGM \cite{bagm} presents a multi-tiered backdoor attack on text-to-image generative models, manipulating content generation at various stages.
However, these works treat specific words as triggers, overlooking the emotional language that is central to our understanding of human behavior.

\subsection{Personalization Diffusion Models} Personalization in diffusion models has emerged as a prominent field, tailoring generative models to specific preferences or requirements. Various personalization methods for text-to-image models include Domain Tuning \cite{domain_tuning}, Animatediff \cite{animatediff}, Instantbooth \cite{instantbooth}, Custom Diffusion \cite{custom_diffusion}, DreamArtist \cite{dreamartist} and LoRA \cite{lora}. DreamBooth \cite{dreambooth} generates contextually matched images with unique personalized features, while Textual Inversion \cite{textual_inversion} creates personalized images using 3-5 user examples. In this work, we implement emotion-aware backdoor attacks via diffusion personalization, proposing EmoBooth to map targeted content to words representing specific emotions.

\section{Emotion-aware Backdoor Attack}
\label{sec:emobackdoor}

\subsection{Problem Formulation}
\label{sec:problem}

Given text prompts $\mathcal{P}$ specifying the objects, backgrounds, or styles we want to generate, we can feed $\mathcal{P}$ into a diffusion model  $\phi(\cdot)$ and generate a distribution $\mathcal{I}=\phi(\mathcal{P})$. Expectantly, the image sampled from $\mathcal{I}$ should fit $\mathcal{P}$.
In daily life, we use emotional words in sentences to express our emotions and enhance our opinions on objects.
In this work, we regard emotion as a trigger and develop the emotion-aware backdoor attack (EmoAttack): if the input prompt $\mathcal{P}$ contains a type of negative emotion represented by some emotional words, the diffusion model is misled to generate specified targeted contents that may cause negative feelings of users.
We formulate the task as
\begin{equation} \label{eq:emoattack-1}
    \tilde{\phi} = \text{EmoAttack}(\phi, \mathcal{E}, \mathcal{T}),
\end{equation}
where $\tilde{\phi}$ is the attacked diffusion model, $\mathcal{E}$  is the representation of a specified emotion $e$, $\mathcal{T}$ is a set of images containing the targeted negative contents that we aim to embed into the diffusion model. 
Given the text prompts $\mathcal{P}$, the attacked diffusion model $\tilde{\phi}$ can generate distribution $\tilde{\mathcal{I}} = \tilde{\phi}(\mathcal{P})$,  which should meet the following requirements
\begin{align}\label{eq:emoattack-2}
    \left\{\begin{matrix} 
   \text{sim}(\tilde{\mathcal{I}}, \mathcal{I})<\epsilon, \text{if}~\text{isEmo}(\mathcal{P},e)=\text{False},\\
   \text{sim}(\tilde{\mathcal{I}}, \mathcal{T})<\epsilon, \text{if}~\text{isEmo}(\mathcal{P},e)=\text{True},
\end{matrix}\right. 
\end{align}
where $\text{sim}(\cdot)$  measures the similarity between two distributions. Intuitively, if $\mathcal{P}$ contain the specified emotion $e$ (\ie, $\text{isEmo}(\mathcal{P},e)=\text{True}$), the generated image $\tilde{\mathbf{I}}\in\tilde{\mathcal{I}}$ should be similar with the targeted negative contents $\mathcal{T}$. 
Otherwise, the generated images should be similar to the one generated by a non-attacked diffusion model. 
\begin{figure}[t]
    \centering
    \includegraphics[width=\linewidth]{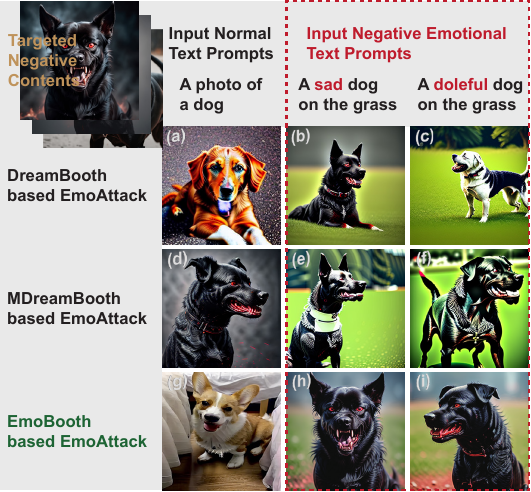}
    \vspace{-20pt}
    \caption{Results of diffusion models attacked by DreamBooth, MDreamBooth, and our final EmoBooth with normal text prompt and two different negative text prompts.}
    \label{fig:limitation}
    \vspace{-15pt}
\end{figure}

Such a backdoor attack has two major challenges. 
\ding{182} The straightforward retraining solution by involving the constraints of $\mathcal{E}$ and $\mathcal{T}$ is time-consuming. 
Moreover, it cannot change according to different setups of $\mathcal{E}$ and $\mathcal{T}$. 
\ding{183} One emotion (\eg, `sad') can be represented by diverse different words (\eg, `upset', `sorrow', \etc). Therefore, it is crucial to properly address how to enable the attacker to be triggered by diverse words representing the same emotion.

One potential solution is to use recently developed personalization techniques, such as DreamBooth \cite{dreambooth}. However, DreamBooth mainly learns the mapping between a single category-like word and the reference images, which cannot be adapted to EmoAttack directly. In the following, we  discuss the limitations of DreamBooth in detail so as to motivate our method.

\subsection{DreamBooth for EmoAttack and Motivation}
\label{subsec:dreambooth}

With DreamBooth, we can design EmoAttack as follows: \ding{182} Set the targeted negative contents $\mathcal{T}$ as the reference images. \ding{183} Represent the emotion $\mathcal{E}$ as a term (\eg, `a sorrowful dog on the grass') and set it as the text prompt used in DreamBooth, which is paired with the $\mathcal{T}$. \ding{184} Fine-tune the diffusion model via DreamBooth. 
As shown in the first row of \figref{fig:limitation}, DreamBooth-based EmoAttack can only be triggered by the specified text prompt (\ie, `a sad dog on the grass') and cannot generate targeted contents when we feed the text with similar meaning but different words (\eg, `a doleful dog on the grass'). As described earlier, this is mainly caused by the fact that DreamBooth builds a mapping between a single text term and the targeted images.

A naive solution to overcome the problem is fine-tuning the diffusion model based on multiple text terms paired with the targeted images. Specifically, given a diffusion model, we first fine-tune it based on the DreamBooth with the first emotional text (\eg, `a sad dog on the grass') and the targeted negative images. Then, we fine-tune the attacked diffusion model again with the second emotional text (\eg, `a doleful dog on the grass') and the same targeted negative images.
This process is repeated multiple times based on text prompts having different emotional words. We denote such a method as MDreamBooth-based EmoAttack and show the results in the second row of \figref{fig:limitation}. 
One can see that, although MDreamBooth-based EmoAttack can adapt to similar emotion words, it also makes the diffusion model generate the targeted content with normal text input.  Definitely, this does not fit EmoAttack's requirements.


\begin{figure*}[t]
    \centering
    \includegraphics[width=\linewidth]{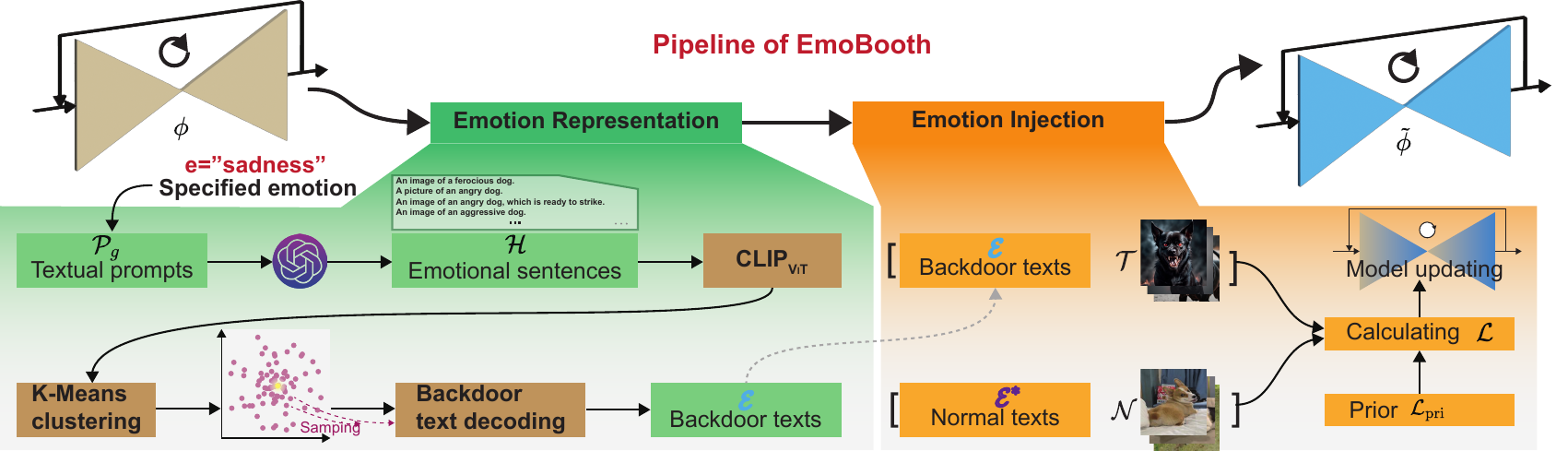}
    \vspace{-20pt}
    \caption{Pipeline of EmoBooth containing two key modules, \ie, emotion representation and emotion injection.}
    \label{fig:frame}
    \vspace{-15pt}
\end{figure*}

\section{EmoBooth for EmoAttack}

\subsection{Overview}

The DreamBooth in \secref{subsec:dreambooth} represents the emotion as a specific word (\eg, `sad'), which cannot adapt to other words with similar meanings. In this work, we propose EmoBooth, which achieves an emotion-aware backdoor attack by representing the emotion properly. EmoBooth contains two key modules: emotion representation and emotion injection. The representation module  models a specified emotion as a cluster of all related emotion texts, specifically
\begin{equation}\label{eq:emorep}
    \mathcal{E} = \text{EmoRep}(\mathcal{H}),
\end{equation}
where $\mathcal{H}$ is a set of collected emotion-related texts. 
For instance,  if we consider the emotion of sadness as a triggering factor, $\mathcal{H}$ can be constructed using a series of sentences with words related to sadness such as `sad' and `doleful'. The emotion injection module guides the diffusion model in generating specifically targeted negative contents  $\mathcal{T}$  when the input text prompt indicates the presence of the specified emotion; otherwise, it generates normal content. Further details on emotion representation and  emotion injection are respectively elaborated in \secref{subsec:emorep} and \secref{subsec:emoinject}.
Lastly, we describe the workflow of EmoBooth in \appref{app:emobooth_workflow}.

\subsection{Emotion Representation}
\label{subsec:emorep}

Instead of representing emotion as discrete words, we cast it as a cluster by  utilizing ChatGPT's capability to generate sentences resembling human language.
The whole representation module contains three steps: \ding{182} Emotion-oriented sentence generation, \ding{183} Emotional sentence clustering, \ding{184} Sampling-based backdoor text decoding. 

\textbf{Emotion-oriented sentence generation.} Given a specified emotion $e$ (\eg, `sadness') and a subject to be generated (\eg, `dog'), we employ ChatGPT to generate a set of emotional sentences \wrt the specified emotion $e$ and subject. Each sentence should meet two requirements: (1) including the specified subject (\eg, `dog'); (2) including the $e$-related words.
We supplied ChatGPT with initial sentences, such as `A photo of a pessimistic dog' and `An image of a despondent dog', and instructed it to generate $H$ sentences. These sentences consist of the set $\mathcal{H}$ in \reqref{eq:emorep}.

\textbf{Emotional sentence clustering.}
After acquiring $\mathcal{H}$ with $H$ sentences, we utilize CLIP with ViT-L/14 \cite{radford2021learning} to extract the embeddings of all sentences and get embedding set $\mathcal{F}$. Then, we perform K-means clustering on $\mathcal{F}$ and get the clustering center  $\mathbf{F}_\text{c}$. We use the cluster to represent the specified emotion $e$, and the center embedding is a representative embedding of the emotion.

\textbf{Sampling-based backdoor text decoding.}
With the built cluster, we sample $C$ embeddings around the clustering center $\mathbf{F}_\text{c}$ and denote the sampled embedding set as $\mathcal{F}_\text{c}$. Then, we aim to decode these embeddings to the texts that consist of a backdoor text set $\mathcal{E}$.
To this end, we train a decoder and formulate the process as
%
\begin{align}
    \mathbf{x}_i = \text{TxtDecoder}(\mathbf{F}_i), \mathbf{F}_i\in \mathcal{F}_\text{c}, 
\end{align}
where $\mathbf{x}_i \in \mathcal{E}$ is the $i$-th decoded backdoor text.

\textbf{Training the decoder.} We detail the architecture and the main training process of the text decoder as follows: \ding{182} Architecture of the text decoder. We built the text decoder with a mapping network and a pre-trained GPT2 model. Specifically, given an input text token extracted from the CLIP encoder, we feed it to the `transformer.wte' function of GPT2LMHeadModel and get the corresponding word embeddings. Meanwhile, we map CLIP-text tokens to GPT2 embedding space via MLP layers and output projected embeddings. Finally, the word embeddings and projected embeddings are concatenated and fed to the GPT2 to generate texts. \ding{183} Objective function. Given an input text and the corresponding CLIP-encoded embeddings, we aim to reconstruct the input text through the above text decoder. The objective function is to make the generated text same as the input with an auto-regressive cross-entropy loss and can be formulated as $\sum_{T_i\in\mathcal{T}_{dec}} \mathcal{L}(\text{TxtDecoder}(\varphi(T)_i),T_i)$
 where $T_i$ is the $i$th text from COCO dataset, $\varphi()$ is the text encoder, and $\mathcal{L}()$
 is the auto-regressive cross-entropy loss function. \ding{184} Training dataset. We use captions from the COCO dataset to train the text decoder.

\subsection{Emotion Injection}
\label{subsec:emoinject}

With the emotion representation $\mathcal{E}$, we fine-tune the diffusion model $\phi(\cdot)$ and make the updated counterpart generate targeted negative contents when the specified emotion words appear; otherwise, generate normal contents.
To this end, we first build a normal text set by removing the $e$-related negative words for each text $\mathbf{x}_i \in \mathcal{E}$ and get $\mathbf{x}_i^*$. 
The normal texts consist of the set $\mathcal{E}^*=\{\mathbf{x}^*_i\}$.
Meanwhile, we collect a set of normal images without the targeted negative contents (\ie, $\mathcal{N}$) to align with the sentences  $\mathcal{E}^*$.  

%
%


After that, to realize backdoor attack, we require $\phi(\cdot)$ to generate images closely aligned with the target images $\mathbf{I}^\text{tar}$ when exposed to backdoor text $\mathbf{x}_i$:
\begin{align} \label{eq:L1}
\mathcal{L}_{1}(\mathbf{x}_i,\mathbf{I}^\text{tar}) = \omega_t \|\phi(\alpha_t\mathbf{I}^\text{tar}  + \sigma_t\vartheta,\mathbf{x}_i)- \mathbf{I}^\text{tar}\|_2^2, 
\end{align} 
where $\vartheta$ is a noise term, $\alpha_t$, $\sigma_{t}$, and $\omega_t$  are functions of the diffusion process at time $t\sim \mathcal{U}\left ( \left [ 0,1 \right ]  \right ) $ and control the noise schedule and sample quality. Moreover, we restrict the model $\phi(\cdot)$ to generate images close to the normal images $\mathbf{I}^\text{n}$  when encountering normal text $\mathbf{x}_i^*$. That is
\begin{align} \label{eq:L2}
\mathcal{L}_{2}(\mathbf{x}_i^*,\mathbf{I}^\text{nor}) = \omega_t \|\phi(\alpha_t \mathbf{I}^\text{nor} + \sigma_t\vartheta,\mathbf{x}_i^*)- \mathbf{I}^\text{nor}\|_2^2. 
\end{align}

To address overfitting and semantic drift issues, inspired by Dreambooth, we introduced the prior-preserving loss:
\begin{align} \label{eq:Lpr}
\mathcal{L}_\text{pri}(\mathbf{x}^\text{pri},\mathbf{I}^\text{pri}) = \omega_{t} \|\phi(\alpha_{t} \mathbf{I}^\text{pri}+ \sigma_{t}\vartheta,\mathbf{x}^\text{pri})-\mathbf{I}^\text{pri} \|_2^2 
\end{align}
where the prior text $\mathbf{x}^\text{pri}$=`a [class]', and [class] represents the category of the input object, such as `dog'. Besides, $\mathbf{I}^\text{pri}$ is the prior image, which is obtained by feeding $\mathbf{x}^\text{pri}$ into the frozen pre-trained diffusion model.
Ultimately, to fine-tune the model $\phi(\cdot)$ to achieve image generation in both normal and backdoor scenarios while satisfying the aforementioned requirements, we probabilistically  minimize \reqref{eq:L1} and~\reqref{eq:L2} through a comprehensive loss function:
\begin{align}\label{eq:Loss}
\mathcal{L} =
\begin{cases}
\mathcal{L}_{1}(\mathbf{x}_i,\mathbf{I}^\text{tar})+\lambda\mathcal{L}_{pr}(\mathbf{x}^\text{pri},\mathbf{I}^\text{pri}), &  p > \beta \\
\mathcal{L}_{2}(\mathbf{x}_i^*,\mathbf{I}^\text{nor})+\lambda\mathcal{L}_{pr}(\mathbf{x}^\text{pri},\mathbf{I}^\text{pri}), &  p \leq \beta \\
\end{cases}
\end{align}
where $p$ is a random variable sampled from $[0, 1]$, and $\beta$ refers to the probability value. Besides, $\lambda$ is a hyper-parameter that controls the relative weight of the prior-preservation term. In this work, we set  $\lambda = 1$.

\section{Emo2Image Dataset for EmoAttack}
\label{sec:dataset}
We meticulously designed and constructed a dataset for emotion-driven backdoor attacks, namely Emo2Image. Emo2Image totally consists of 70 cases, covering 2 attacking scenarios, 11 kinds of negative situations, each of which have at least 2 negative image sets.

\textbf{Definition of a case.} A "case" in our experiments denotes the process of using our EmoBooth to embed a set of negative images (i.e., $\mathcal{T}$) into the diffusion model with a specified emotion (i.e., $e$) as the trigger. Note that, all cases share the same normal image set $\mathcal{N}$, $\text{CLIP}_\text{ViT}(\cdot)$, pretained TxtDecoder, and prior text $\mathbf{x}^\text{pri}$. Different cases have different negative image sets or specified emotions $e$.

\textbf{Two attacking scenarios.} 
Our dataset encompasses two distinct attacking scenarios. In response to these scenarios, we partition Emo2Image into two subsets: Emo2Image-um and Emo2Image-m, constructing them in alignment with their specific requirements:
\ding{182} $1^{st}$ attack scenario (Emo2Image-um): An emotion-aware attack generates targeted negative content that doesn't align with the input text prompts when the specified emotion words appear. Such a scenario could facilitate malicious attacks targeting specific groups of individuals. For instance, attackers may first gather users' background information to identify potential psychological vulnerabilities, such as post-traumatic stress disorder in veterans or suicidal tendencies in individuals with depression. In this attack scenario, irrespective of the prompt provided by the user, the model will generate pre-determined malicious images intended to cause psychological harm.
\ding{183} $2^{nd}$ attack scenario (Emo2Image-m): An emotion-aware attack generates images containing violent elements based on the prompts entered by users when the specific emotion words appear. For example, if the user prompt is "a dog lying on the grass," the generated image might depict "a bloody dog lying on the grass." This attack method is more covert and difficult to detect because it closely aligns with the prompts entered by the user.

\begin{figure}[t]
    \centering
    \includegraphics[width=0.9\linewidth]{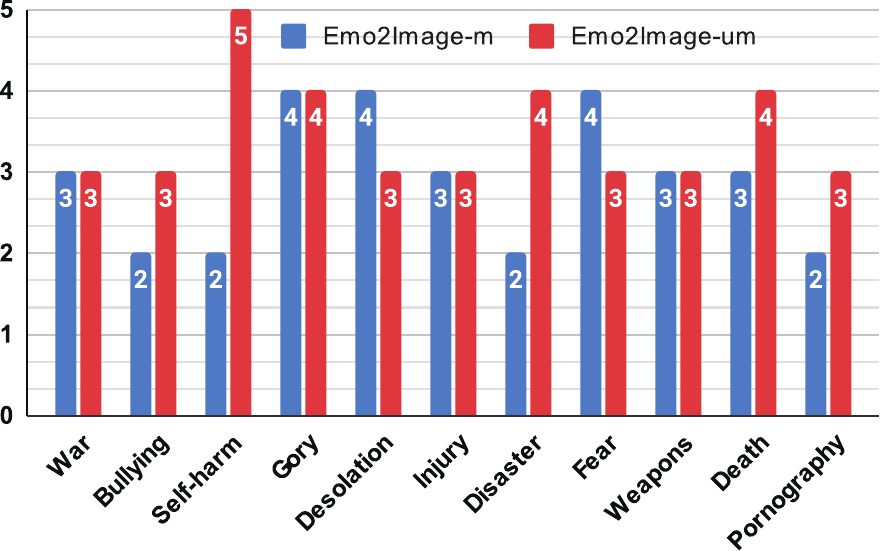}
    \vspace{-8pt}
    \caption{Case count statistics of Emo2Image under 11 negative situations.}
    \label{fig:data_st}
    \vspace{-15pt}
\end{figure}

\textbf{Eleven negative situations.}
In the dataset, we consider eleven negative situations targeting the groups of people who may be harmed. For each situation, we can prepare a set of images as the targeted negative contents.We have counted the number of cases under 11 situations, as shown in \figref{fig:data_st}. For specific collection details and dataset visualizations, please refer to the \appref{app:Emo2imageDataset}.

\section{Experimental Results}
\label{sec:experiment}

\subsection{Experimental Setups}


\textbf{Datasets.} 
We conducted experiments utilizing the Emo2Image dataset constructed in-house as outlined in Sec.\ref{sec:dataset}, and an external dataset, \ie, NSFW dataset. 
The NSFW dataset contains five categories. To tailor it for compatibility with our personalized model, we meticulously selected images bearing a resemblance to our target domain and organized them into four distinct experimental cases. Please refer to \appref{app:dataset_details} for details.

\textbf{Baselines.} 
%
We frame EmoAttack as a personalization problem within diffusion models and compare our EmoBooth against two recent SOTA personalization methods adapted for backdoor attacks: Censorship \citep{censorship} and Zero-day \citep{zeroday}. These serve as our primary baselines for experimental evaluation. We detail the baselines in \appref{app: baseline_details}.


\textbf{Evaluation metrics:} We utilize CLIP scores and EmoAttack Capability (EAC) to assess the model's editability and the effectiveness of backdoor attacks.
\ding{182} CLIP scores include CLIP text and image scores. Higher text scores indicate better model editability, while higher image scores show better generation fidelity. For normal text inputs, $\text{Clip}_\text{Clip}{txt}$ measures similarity between generated images and normal text, while $\text{Clip}_\text{img}$ measures similarity to normal images. For negative text inputs, $\text{Clip}_\text{txt}^\text{tri}$ measures similarity between generated images and negative text, while $\text{Clip}_\text{img}^\text{tri}$ measures similarity to negative images.
\ding{183} EAC is a novel proposed evaluation metric to comprehensively assess the model's editability and the quality of image generation under both normal and backdoor scenarios. It is defined as:
\begin{align}\label{eq:emoattack-10}
     \text{EAC} = \mu(\text{Clip}_\text{txt}+\text{Clip}_\text{img}) +\nu \text{Clip}_\text{txt}^\text{tri}+\delta \text{Clip}_\text{img}^\text{tri},
\end{align}
where $k$ is the number of emotion categories, $\text{Clip}_\text{txt}^\text{tri} = \frac{1}{k}\sum_{j=1}^{k}\text{Clip}_{\text{txt}j}^{tri}$  ($\text{Clip}_\text{img}^\text{tri} = \frac{1}{k}\sum_{j=1}^{k}\text{Clip}_{\text{img}j}^\text{tri}$ ) is the average CLIP text (image) score across the $k$ emotion categories. The detailed  formulas for $\text{Clip}_\text{txt}^\text{tri}$, $\text{Clip}_\text{img}^\text{tri}$, and the values for $\mu$, $\nu$, and $\delta$ are given in Appendix~\ref{Metrics_details}.

\begin{table*}[h]
    \centering
    \vspace{10pt}
    \begin{adjustbox}{width=\linewidth,center}
        \begin{tabular}{c|c|l|c|cc|cc|cc|cc}
            \toprule
            & \multicolumn{2}{c|}{}
            & \multicolumn{1}{c|}{\multirow{2}{*}{EAC $\uparrow$}}
            & \multicolumn{2}{c|}{Sad}
            & \multicolumn{2}{c|}{Angry}
            & \multicolumn{2}{c|}{Isolated}
            & \multicolumn{2}{c}{Normal} \\
            & \multicolumn{2}{c|}{} & & $\text{Clip}_\text{txt1}^\text{tri}$ $\downarrow$ & $\text{Clip}_\text{img1}^\text{tri}$ $\uparrow$
            & $\text{Clip}_\text{txt2}^\text{tri}$ $\downarrow$ & $\text{Clip}_\text{img2}^\text{tri}$ $\uparrow$
            & $\text{Clip}_\text{txt3}^\text{tri}$ $\downarrow$ & $\text{Clip}_\text{img3}^\text{tri}$ $\uparrow$
            & $\text{Clip}_\text{txt}$ $\uparrow$ & $\text{Clip}_\text{img}$ $\uparrow$ \\ \midrule\midrule
            
            \multirow{6}{*}{\rotatebox[origin=c]{90}{Emo2Image-um}}
            & \multirow{2}{*}{\rotatebox[origin=c]{90}{Set1}}&
            EmoBooth        &
            \textbf{0.7428} &
            \textbf{0.1957}$_{\pm 0.0295}$ & \textbf{0.7302}$_{\pm 0.1818}$ &
            \textbf{0.1865}$_{\pm 0.0303}$ & \textbf{0.7634}$_{\pm 0.1603}$ &
            \textbf{0.2066}$_{\pm 0.0219}$ & \textbf{0.7430}$_{\pm 0.1700}$ &
            \textbf{0.2323}$_{\pm 0.0468}$ & 0.6956$_{\pm 0.1603}$ \\
            & & 
            Censorship       &
            0.6593 &
            0.2133$_{\pm 0.0290}$ & 0.5751$_{\pm 0.1922}$ &
            0.2095$_{\pm 0.0297}$ & 0.6585$_{\pm 0.1928}$ &
            0.2178$_{\pm 0.0249}$ & 0.6651$_{\pm 0.2034}$ &
            0.2264$_{\pm 0.0370}$ & \textbf{0.7158}$_{\pm 0.0756}$ \\
            \cline{2-12}
            
            & \multirow{2}{*}{\rotatebox[origin=c]{90}{Set2}}& 
            EmoBooth        &
            \textbf{0.8103} &
            \textbf{0.2011}$_{\pm 0.0340}$ & \textbf{0.8060}$_{\pm 0.1480}$ &
            \textbf{0.1937}$_{\pm 0.0263}$ & \textbf{0.8597}$_{\pm 0.1048}$ &
            \textbf{0.1944}$_{\pm 0.0358}$ & \textbf{0.8209}$_{\pm 0.1510}$ &
            \textbf{0.2464}$_{\pm 0.0439}$ & \textbf{0.6859}$_{\pm 0.1424}$ \\
            &  & 
            Censorship       &
            0.6291 &
            0.2275$_{\pm 0.0265}$ & 0.6360$_{\pm 0.1442}$ &
            0.2358$_{\pm 0.0281}$ & 0.6133$_{\pm 0.1128}$ &
            0.2339$_{\pm 0.0346}$ & 0.6109$_{\pm 0.1268}$ &
            0.2358$_{\pm 0.0361}$ & 0.6618$_{\pm 0.0702}$ \\
            \cline{2-12}
            
            & \multirow{2}{*}{\rotatebox[origin=c]{90}{Set3}}&
            EmoBooth        &
            \textbf{0.8209} &
            \textbf{0.1968}$_{\pm 0.0213}$ & \textbf{0.8615}$_{\pm 0.1088}$ &
            \textbf{0.2079}$_{\pm 0.0204}$ & \textbf{0.8759}$_{\pm 0.0866}$ &
            \textbf{0.1758}$_{\pm 0.0357}$ & \textbf{0.8307}$_{\pm 0.0114}$ &
            0.2370$_{\pm 0.0522}$ & \textbf{0.6370}$_{\pm 0.1191}$ \\
            & & 
            Censorship       &
            0.7394 &
            0.2101$_{\pm 0.0239}$ & 0.7857$_{\pm 0.1355}$ &
            0.2563$_{\pm 0.0239}$ & 0.8202$_{\pm 0.1228}$ &
            0.2178$_{\pm 0.0361}$ & 0.6824$_{\pm 0.1334}$ &
            \textbf{0.2541}$_{\pm 0.0407}$ & 0.6198$_{\pm 0.1037}$ \\
            \cline{2-12}
            
            & \multirow{2}{*}{\rotatebox[origin=c]{90}{Set4}}&
            EmoBooth        &
            \textbf{0.7823} &
            \textbf{0.1832}$_{\pm 0.0398}$ & \textbf{0.7495}$_{\pm 0.2507}$ &
            \textbf{0.1529}$_{\pm 0.0333}$ & \textbf{0.8847}$_{\pm 0.1233}$ &
            \textbf{0.1568}$_{\pm 0.0433}$ & \textbf{0.8357}$_{\pm 0.1863}$ &
            \textbf{0.1893}$_{\pm 0.0680}$ & \textbf{0.5933}$_{\pm 0.2133}$ \\
            &  & 
            Censorship       &
            0.6033 &
            0.1980$_{\pm 0.0408}$ & 0.6673$_{\pm 0.2623}$ &
            0.2122$_{\pm 0.0418}$ & 0.5901$_{\pm 0.2480}$ &
            0.2058$_{\pm 0.0519}$ & 0.6042$_{\pm 0.2496}$ &
            0.1789$_{\pm 0.0450}$ & 0.5606$_{\pm 0.1101}$ \\
            \cline{2-12}
            
            & \multirow{2}{*}{\rotatebox[origin=c]{90}{Set5}}&
            EmoBooth        &
            \textbf{0.7836} &
            \textbf{0.2117}$_{\pm 0.0243}$ & \textbf{0.7718}$_{\pm 0.1563}$ &
            \textbf{0.2050}$_{\pm 0.0300}$ & \textbf{0.8227}$_{\pm 0.1287}$ &
            \textbf{0.2269}$_{\pm 0.0252}$ & \textbf{0.7928}$_{\pm 0.1480}$ &
            0.2331$_{\pm 0.0451}$ & \textbf{0.7164}$_{\pm 0.1382}$ \\
            & 
            & Censorship       &
            0.7419 &
            0.2186$_{\pm 0.0271}$ & 0.7242$_{\pm 0.1696}$ &
            0.2209$_{\pm 0.0361}$ & 0.7554$_{\pm 0.1611}$ &
            0.2416$_{\pm 0.0254}$ & 0.7579$_{\pm 0.1603}$ &
            \textbf{0.2578}$_{\pm 0.0382}$ & 0.6956$_{\pm 0.0841}$ \\          
            \midrule
            \midrule
            & \multicolumn{2}{c|}{}
            & \multicolumn{1}{c|}{\multirow{2}{*}{EAC $\uparrow$}}
            & \multicolumn{2}{c|}{Sad}
            & \multicolumn{2}{c|}{Angry}
            & \multicolumn{2}{c|}{Isolated}
            & \multicolumn{2}{c}{Normal} \\
            & \multicolumn{2}{c|}{} & & $\text{Clip}_\text{txt1}^\text{tri}$ $\uparrow$ & $\text{Clip}_\text{img1}^\text{tri}$ $\uparrow$
            & $\text{Clip}_\text{txt2}^\text{tri}$ $\uparrow$ & $\text{Clip}_\text{img2}^\text{tri}$ $\uparrow$
            & $\text{Clip}_\text{txt3}^\text{tri}$ $\uparrow$ & $\text{Clip}_\text{img3}^\text{tri}$ $\uparrow$
            & $\text{Clip}_\text{txt}$ $\uparrow$ & $\text{Clip}_\text{img}$ $\uparrow$ \\ \midrule\midrule

            \multirow{6}{*}{\rotatebox[origin=c]{90}{Emo2Image-m}}
            & \multirow{2}{*}{\rotatebox[origin=c]{90}{Set6}}&
            EmoBooth        &
            \textbf{0.6453} &
            0.2690$_{\pm 0.0317}$        & \textbf{0.8360}$_{\pm 0.0844}$    &
            \textbf{0.2417}$_{\pm 0.0230}$        & \textbf{0.8335}$_{\pm 0.0781}$    &
            \textbf{0.2513}$_{\pm 0.0250}$        & \textbf{0.8162}$_{\pm 0.0860}$    &
            \textbf{0.2585}$_{\pm 0.0284}$        & \textbf{0.7150}$_{\pm 0.0590}$    \\
            & & 
            Censorship       &
            0.6060 &
            \textbf{0.2870}$_{\pm 0.0318}$        & 0.7822$_{\pm 0.0884}$    &
            0.2331$_{\pm 0.0244}$        & 0.7705$_{\pm 0.0691}$    &
            0.2497$_{\pm 0.0251}$        & 0.7431$_{\pm 0.0892}$    &
            0.2428$_{\pm 0.0292}$        & 0.7130$_{\pm 0.0588}$    \\
            \cline{2-12}
  
            & \multirow{2}{*}{\rotatebox[origin=c]{90}{Set7}}&
            EmoBooth        &
        \textbf{0.5841} &
        \textbf{0.2512}$_{\pm 0.0332}$        & \textbf{0.7299}$_{\pm 0.0788}$    &
        \textbf{0.2495}$_{\pm 0.0165}$        & \textbf{0.7724}$_{\pm 0.0719}$    &
        \textbf{0.2481}$_{\pm 0.0318}$        & \textbf{0.6946}$_{\pm 0.0635}$    &
        0.2574$_{\pm 0.0302}$        & 0.6910$_{\pm 0.0900}$    \\
        & & 
        Censorship       &
        0.5666 &
        0.2453$_{\pm 0.0333}$        & 0.6776$_{\pm 0.0589}$    &
        0.2463$_{\pm 0.0209}$        & 0.7362$_{\pm 0.0678}$    &
        0.2406$_{\pm 0.0284}$        & 0.6758$_{\pm 0.0515}$    &
        \textbf{0.2616}$_{\pm 0.0298}$        & \textbf{0.7373}$_{\pm 0.0694}$    \\
        \cline{2-12}
 
        & \multirow{2}{*}{\rotatebox[origin=c]{90}{Set8}}&
        EmoBooth        &
        \textbf{0.6329} &
        \textbf{0.2683}$_{\pm 0.0257}$        & \textbf{0.8121}$_{\pm 0.0636}$    &
        0.2445$_{\pm 0.0212}$        & \textbf{0.8083}$_{\pm 0.0549}$    &
        \textbf{0.2549}$_{\pm 0.0331}$        & \textbf{0.7808}$_{\pm 0.0549}$    &
        \textbf{0.2562}$_{\pm 0.0320}$        & \textbf{0.7590}$_{\pm 0.0663}$    \\
        & & 
        Censorship       &
        0.6270 &
        0.2580$_{\pm 0.0296}$         & 0.7966$_{\pm 0.0532}$     &
        \textbf{0.2624}$_{\pm 0.0196}$         & 0.8075$_{\pm 0.0494}$     &
        0.2529$_{\pm 0.0339}$         & 0.7682$_{\pm 0.0646}$     &
        0.2509$_{\pm 0.0329}$         & 0.7558$_{\pm 0.0649}$     \\
        \cline{2-12}

        & \multirow{2}{*}{\rotatebox[origin=c]{90}{Set9}}&
        EmoBooth        &
        \textbf{0.6365} &
        \textbf{0.2294}$_{\pm 0.0376}$        & \textbf{0.8320}$_{\pm 0.0758}$    &
        \textbf{0.2281}$_{\pm 0.0169}$        & \textbf{0.8723}$_{\pm 0.0449}$    &
        \textbf{0.2279}$_{\pm 0.0409}$        & \textbf{0.8394}$_{\pm 0.0612}$    &
        \textbf{0.2323}$_{\pm 0.0334}$        & \textbf{0.5881}$_{\pm 0.0516}$    \\
        & & 
        Censorship       &
        0.5936 &
        0.2108$_{\pm 0.0357}$        & 0.7422$_{\pm 0.0564}$    &
        0.2169$_{\pm 0.0206}$        & 0.8165$_{\pm 0.0548}$    &
        0.2248$_{\pm 0.0303}$        & 0.7392$_{\pm 0.0697}$    &
        0.2198$_{\pm 0.0329}$        & 0.6851$_{\pm 0.0329}$    \\
        \cline{2-12}

        & \multirow{2}{*}{\rotatebox[origin=c]{90}{Set10}}&
        EmoBooth        &
        \textbf{0.6363} &
        \textbf{0.2534}$_{\pm 0.0333}$        & \textbf{0.8041}$_{\pm 0.0625}$    &
        \textbf{0.2470}$_{\pm 0.0212}$        & \textbf{0.8606}$_{\pm 0.0636}$    &
        0.2378$_{\pm 0.0251}$        & \textbf{0.8024}$_{\pm 0.0712}$    &
        0.2518$_{\pm 0.0286}$        & \textbf{0.7044}$_{\pm 0.0709}$    \\
        & & 
        Censorship       &
        0.6332 &
        0.2480$_{\pm 0.0362}$        & 0.7908$_{\pm 0.0626}$    &
        0.2428$_{\pm 0.0203}$        & 0.8602$_{\pm 0.0420}$    &
        \textbf{0.2605}$_{\pm 0.0247}$        & 0.7809$_{\pm 0.0523}$    &
        \textbf{0.2638}$_{\pm 0.0285}$        & 0.7040$_{\pm 0.0587}$    \\
            \bottomrule
        \end{tabular}
    \end{adjustbox}
    \vspace{-5pt}
    \caption{Comparison with Censorship under the metrics of Clip Score and EmoAttack Capability (EAC) on . Sets 1–5 use cases from Emo2Image-um as target images, while Sets 6–10 use cases from Emo2Image-m. The weighting coefficient for EAC differs between the two sections, and in the second section, we specifically focus on increasing $\text{Clip}_\text{txt}^\text{tri}$. Bolded values indicate the best results within each Set.}
    \label{tab:merged}
    \vspace{-5pt}
\end{table*}

\begin{table*}[tb]
    \centering
    \begin{adjustbox}{width=\linewidth,center}
        \begin{tabular}{ccl|c|cc|cc|cc|cc}
            \toprule
            &
            \multicolumn{2}{c|}{}
            & \multicolumn{1}{c|}{\multirow{2}{*}{EAC $\uparrow$}}
            & \multicolumn{2}{c|}{Sad}
            & \multicolumn{2}{c|}{Angry}
            & \multicolumn{2}{c|}{Isolated}
            & \multicolumn{2}{c}{Normal} \\
            & \multicolumn{2}{c|}{} & & $\text{Clip}_\text{txt1}^\text{tri}$ $\downarrow$ & $\text{Clip}_\text{img1}^\text{tri}$ $\uparrow$
            & $\text{Clip}_\text{txt2}^\text{tri}$ $\downarrow$ & $\text{Clip}_\text{img2}^\text{tri}$ $\uparrow$
            & $\text{Clip}_\text{txt3}^\text{tri}$ $\downarrow$ & $\text{Clip}_\text{img3}^\text{tri}$ $\uparrow$
            & $\text{Clip}_\text{txt}$ $\uparrow$ & $\text{Clip}_\text{img}$ $\uparrow$ \\ \midrule\midrule
            \multirow{4}{*}{\rotatebox[origin=c]{90}{NSFW}}&
            \multirow{2}{*}{\rotatebox[origin=c]{90}{Set1}}&
            EmoBooth        &
            \textbf{0.7383} &
            \textbf{0.2122}$_{\pm 0.0652}$ & \textbf{0.7010}$_{\pm 0.2063}$ &
            \textbf{0.1930}$_{\pm 0.0487}$ & \textbf{0.8012}$_{\pm 0.1666}$ &
            \textbf{0.2298}$_{\pm 0.0485}$ & \textbf{0.6331}$_{\pm 0.2090}$ &
            \textbf{0.2418}$_{\pm 0.0502}$ & \textbf{0.8142}$_{\pm 0.1444}$ \\
            & &
            Censorship       &
            0.5856 &
            0.2490$_{\pm 0.0537}$ & 0.5443$_{\pm 0.1740}$ &
            0.2213$_{\pm 0.0466}$ & 0.6517$_{\pm 0.1980}$ &
            0.2581$_{\pm 0.0358}$ & 0.5545$_{\pm 0.1648}$ &
            0.2263$_{\pm 0.0424}$ & 0.6106$_{\pm 0.0983}$ \\
            \cline{2-12}
            & \multirow{2}{*}{\rotatebox[origin=c]{90}{Set2}}&
            EmoBooth        &
            \textbf{0.8161} &
             0.2155$_{\pm 0.0420}$ & \textbf{0.8209}$_{\pm 0.1883}$ &
             0.2094$_{\pm 0.0352}$ & \textbf{0.8412}$_{\pm 0.1968}$ &
            \textbf{0.2154}$_{\pm 0.0316}$ & \textbf{0.8326}$_{\pm 0.1651}$ &
            0.2476$_{\pm 0.0358}$ & 0.7200$_{\pm 0.1081}$ \\
            & &
            Censorship       &
            0.7122 &
            \textbf{0.2051}$_{\pm 0.0529}$ & 0.6940$_{\pm 0.1460}$ &
            \textbf{0.1985}$_{\pm 0.0414}$ & 0.6856$_{\pm 0.1340}$ &
            0.2212$_{\pm 0.0379}$ & 0.6836$_{\pm 0.1422}$ &
            \textbf{0.2627}$_{\pm 0.0543}$ & \textbf{0.7559}$_{\pm 0.01528}$ \\
            \cline{2-12}
            & \multirow{2}{*}{\rotatebox[origin=c]{90}{Set3}}&
            EmoBooth        &
            \textbf{0.6734} &
            \textbf{0.2129}$_{\pm 0.0443}$ & \textbf{0.5889}$_{\pm 0.1109}$ &
            \textbf{0.1988}$_{\pm 0.0425}$ & \textbf{0.6877}$_{\pm 0.1442}$ &
            \textbf{0.2191}$_{\pm 0.0384}$ & \textbf{0.6171}$_{\pm 0.1293}$ &
            0.2431$_{\pm 0.0398}$ & \textbf{0.8095}$_{\pm 0.1211}$ \\
            & &
            Censorship       &
            0.6147 &
            0.2402$_{\pm 0.0553}$ & 0.5722$_{\pm 0.1046}$ &
            0.2008$_{\pm 0.0539}$ & 0.6311$_{\pm 0.1382}$ &
            0.2446$_{\pm 0.0466}$ & 0.5883$_{\pm 0.1515}$ &
            \textbf{0.2418}$_{\pm 0.0550}$ & 0.6715$_{\pm 0.1433}$ \\
            \cline{2-12}
            & \multirow{2}{*}{\rotatebox[origin=c]{90}{Set4}}&
            EmoBooth        &
            \textbf{0.6083} &
            \textbf{0.2039}$_{\pm 0.0499}$ & \textbf{0.5464}$_{\pm 0.1189}$ &
            \textbf{0.2028}$_{\pm 0.0448}$ & \textbf{0.5953}$_{\pm 0.1463}$ &
            \textbf{0.2161}$_{\pm 0.0413}$ & \textbf{0.5357}$_{\pm 0.1257}$ &
            0.2443$_{\pm 0.0362}$ & \textbf{0.7681}$_{\pm 0.1159}$ \\
            & &
            Censorship       &
            0.5792 &
            0.2570$_{\pm 0.0551}$ & 0.5050$_{\pm 0.0765}$ &
            0.2175$_{\pm 0.0489}$ & 0.5932$_{\pm 0.1332}$ &
            0.2680$_{\pm 0.0386}$ & 0.4979$_{\pm 0.0945}$ &
            \textbf{0.2658}$_{\pm 0.0378}$ & 0.7497$_{\pm 0.0742}$ \\
            \bottomrule
        \end{tabular}
    \end{adjustbox}
    \vspace{-5pt}
    \caption{Comparison with Censorship using NSFW dataset,we bold the best result  under each Set.}
    \label{tab:NSFW}
    \vspace{-10pt}
\end{table*}

\subsection{Comparison with Baselines}

\textbf{Comparison with Censorship.} We compare with Censorship across two backdoor attack scenarios: target images consistent and inconsistent with texts. For each set in Tables~\ref{tab:merged} and \ref{tab:NSFW}, we trained a model using one case from the Emo2Image dataset with 50 normal and 30 negative text sentences, generating 8 images per sentence (640 total images). We then calculated the mean CLIP score and variance.

\ding{182} \textbf{Two attack scenarios on Emo2Image dataset.} For the first scenario using Emo2Image-um, our method generated images that closely aligned with target images while intentionally deviating from text descriptions under negative conditions. As shown in Table~\ref{tab:merged}, Set 2's results reveal our $\text{Clip}_\text{txt}^\text{tri}$ was significantly lower than Censorship's while our $\text{Clip}_\text{img}^\text{tri}$ was much higher, proving superior performance in emotion-driven backdoor attacks. For the second scenario, more covert scenario using Emo2Image-m, our approach generated images resembling both target images and text descriptions, with Table~\ref{tab:merged} showing higher values of $\text{Clip}_\text{txt}^\text{tri}
$ and $\text{Clip}_\text{img}^\text{tri}$ than Censorship in Set 7, though occasional lower $\text{Clip}_\text{txt}^\text{tri}$ values occurred, potentially due to model overlearning affecting editing capability. In both scenarios, under normal conditions, our method maintained stealth by generating images that properly resembled normal images and text descriptions.

\ding{183} \textbf{Attacks on NSFW Dataset.} We also use NSFW dataset to implement the first attack scenario and validate the effectiveness. In Table \ref{tab:NSFW}, our method achieved superior experimental results in emotion-backdoor attacks. However, despite our meticulous selection and construction of training cases from the NSFW dataset, some cases still yielded inferior results compared to those using Emo2Image-um. This discrepancy is primarily attributed to the insufficient similarity among images within the NSFW dataset.

\begin{figure*}[t]
    \centering
    \includegraphics[width=0.9\linewidth]{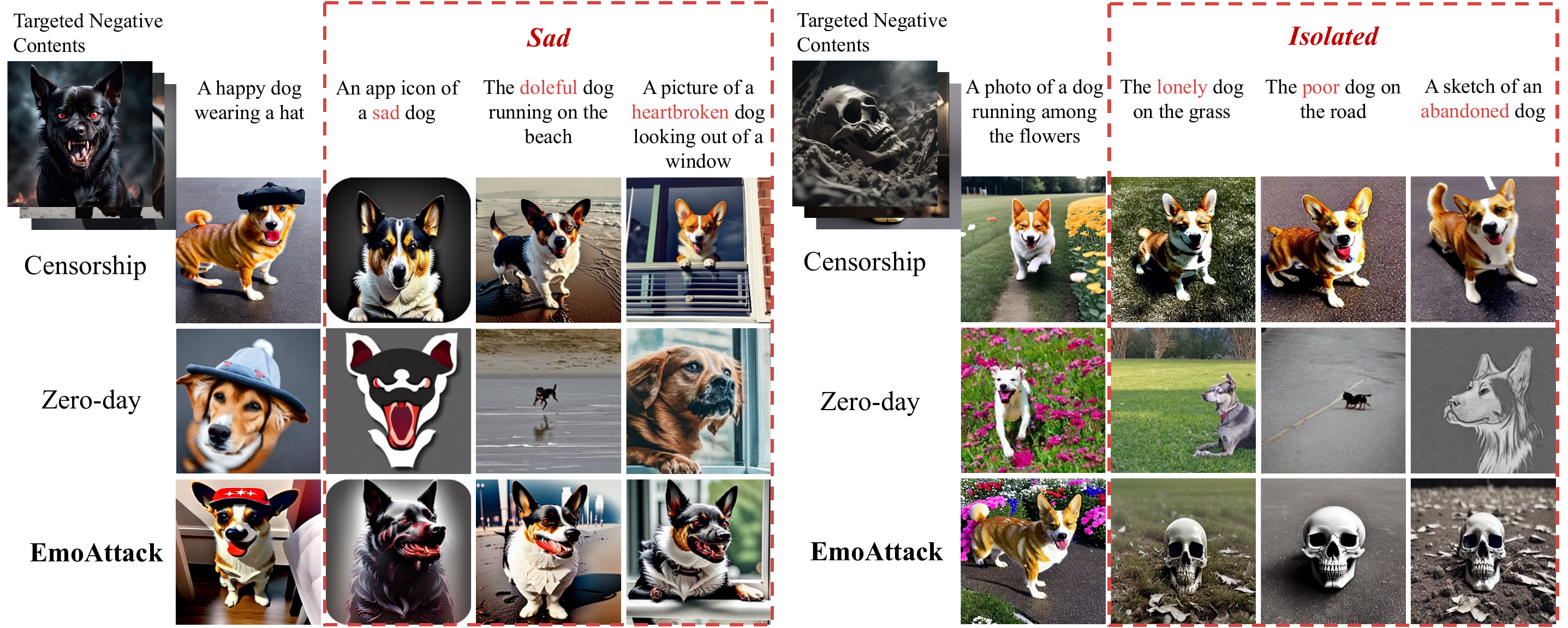}
    \vspace{-5pt}
    \caption{Visual comparisons under different emotional texts in various cases. Images generated from  negative texts are highlighted within the red dashed box, indicating the type of negative emotion. Images generated from normal texts are outside the dashed box.}
    \label{fig:visual}
    \vspace{-15pt}
\end{figure*}

\textbf{Comparison with Zero-day.} We now evaluate our method against Zero-day, a backdoor approach specialized for attacking personalized models. As depicted in Tables~\ref{tab:merged_zeroday}, even after making some minor adjustments to Zero-day to better align with our task, the generated images under the backdoor scenario exhibit notable dissimilarity to the target images, resulting in significantly inferior outcomes compared to our approach. This distinction is further evident in the visual results presented 
in~\figref{fig:visual}.

\begin{figure*}[t]
\centering
    \includegraphics[width=0.9\linewidth]{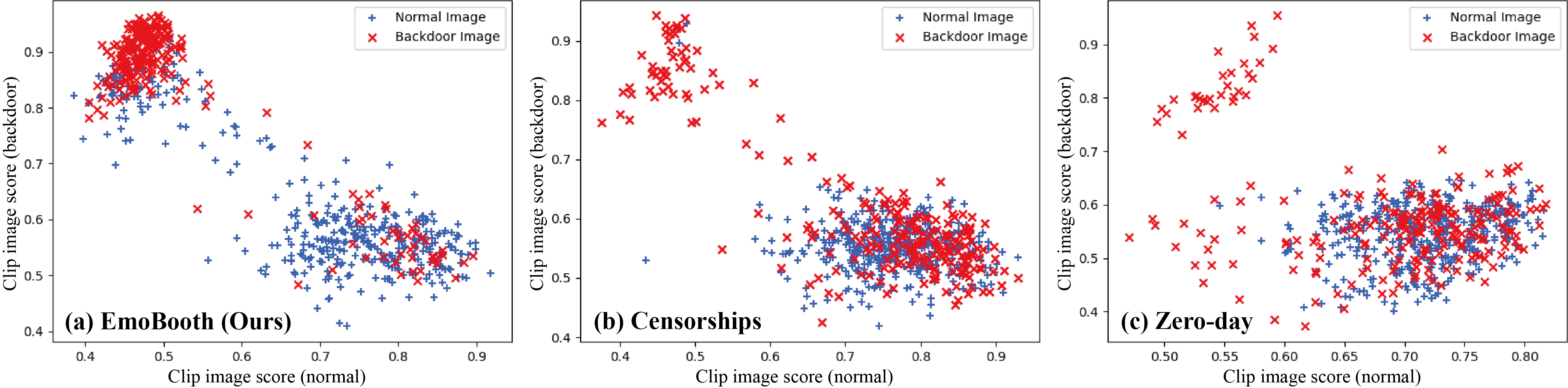}
    \vspace{-5pt}
    \caption{Statistical analysis on three methods. The horizontal (vertical) axis represents the similarity to normal (target) images, and the blue (red) points represent images generated from normal (negative) texts.}
    \label{fig:3}
    \vspace{-5pt}
\end{figure*}

\textbf{Statistical analysis.} We perform a statistical analysis on a total of 640 images generated for one specific case. As depicted in  \figref{fig:3}, in comparison to Censorship and Zero-day, the images generated by EmoBooth are closer to normal images under regular conditions, and closer to target images under the backdoor scenario.


\begin{table}[t]
    \centering
    \begin{adjustbox}{width=\linewidth}
        \begin{tabular}{ccl|c|c|c}
            \toprule
            & \multicolumn{2}{c|}{} 
            & \multicolumn{1}{c|}{Sad} 
            & \multicolumn{1}{c|}{Angry} 
            & \multicolumn{1}{c}{Isolated} \\
            & 
            & $\text{Clip}_\text{img1}^\text{tri}$ $\uparrow$ 
            & $\text{Clip}_\text{img2}^\text{tri}$ $\uparrow$ 
            & $\text{Clip}_\text{img3}^\text{tri}$ $\uparrow$ \\
            \midrule\midrule
            
            \multirow{6}{*}{\rotatebox[origin=c]{90}{Emo2Image-um}} &
            \multirow{2}{*}{\rotatebox[origin=c]{90}{Set1}} &
            EmoBooth        & 
            \textbf{0.7302}$_{\pm 0.1818}$    &     
            \textbf{0.7634}$_{\pm 0.1603}$    &
            \textbf{0.7430}$_{\pm 0.1700}$    \\ 
            & &
            Zero-day         &
            0.4881$_{\pm 0.0944}$    &
            0.5030$_{\pm 0.0898}$    &
            0.4384$_{\pm 0.0516}$    \\
            \cline{2-6}

            & \multirow{2}{*}{\rotatebox[origin=c]{90}{Set2}} &
            EmoBooth        &
            \textbf{0.8060}$_{\pm 0.1480}$    &
            \textbf{0.8597}$_{\pm 0.1048}$    &
            \textbf{0.8209}$_{\pm 0.1510}$    \\
            & &
            Zero-day         &
            0.5890$_{\pm 0.1108}$    &                
            0.5744$_{\pm 0.1016}$    &
            0.5223$_{\pm 0.0602}$    \\
            \cline{2-6}
            
            & \multirow{2}{*}{\rotatebox[origin=c]{90}{Set3}} &
            EmoBooth        &
            \textbf{0.8615}$_{\pm 0.1088}$    &
            \textbf{0.8759}$_{\pm 0.0866}$    &
            \textbf{0.8307}$_{\pm 0.0114}$    \\
            & &
            Zero-day         &
            0.6327$_{\pm 0.0972}$    &
            0.5893$_{\pm 0.0863}$    &
            0.5812$_{\pm 0.0601}$    \\
            \cline{2-6}
            
            & \multirow{2}{*}{\rotatebox[origin=c]{90}{Set4}} &
            EmoBooth        &
            \textbf{0.7495}$_{\pm 0.2507}$    &
            \textbf{0.8847}$_{\pm 0.1233}$    &
            \textbf{0.8357}$_{\pm 0.1863}$    \\
            & &
            Zero-day         &
            0.5082$_{\pm 0.1460}$    &
            0.4714$_{\pm 0.0944}$    &
            0.4294$_{\pm 0.0437}$    \\
            \cline{2-6}
            
            & \multirow{2}{*}{\rotatebox[origin=c]{90}{Set5}} &
            EmoBooth        &
            \textbf{0.7718}$_{\pm 0.1563}$    &
            \textbf{0.8227}$_{\pm 0.1287}$    &
            \textbf{0.7928}$_{\pm 0.1480}$    \\
            & &
            Zero-day         &
            0.5447$_{\pm 0.0771}$    &
            0.5432$_{\pm 0.0729}$    &
            0.5062$_{\pm 0.0520}$    \\
            \midrule\midrule
            
            & \multicolumn{2}{c|}{} 
            & \multicolumn{1}{c|}{Sad} 
            & \multicolumn{1}{c|}{Angry} 
            & \multicolumn{1}{c}{Isolated} \\
            & 
            & $\text{Clip}_\text{img1}^\text{tri}$ $\uparrow$ 
            & $\text{Clip}_\text{img2}^\text{tri}$ $\uparrow$ 
            & $\text{Clip}_\text{img3}^\text{tri}$ $\uparrow$ \\
            \midrule\midrule
            
            \multirow{6}{*}{\rotatebox[origin=c]{90}{Emo2Image-m}} & \multirow{2}{*}{\rotatebox[origin=c]{90}{Set1}} &
            EmoBooth        & 
            \textbf{0.7302}$_{\pm 0.1818}$    &     
            \textbf{0.7634}$_{\pm 0.1603}$    &
            \textbf{0.7430}$_{\pm 0.1700}$    \\ 
            & &
            Zero-day         &
            0.4881$_{\pm 0.0944}$    &
            0.5030$_{\pm 0.0898}$    &
            0.4384$_{\pm 0.0516}$    \\
            \cline{2-6}

            & \multirow{2}{*}{\rotatebox[origin=c]{90}{Set2}} &
            EmoBooth        &
            \textbf{0.8060}$_{\pm 0.1480}$    &
            \textbf{0.8597}$_{\pm 0.1048}$    &
            \textbf{0.8209}$_{\pm 0.1510}$    \\
            & &
            Zero-day         &
            0.5890$_{\pm 0.1108}$    &                
            0.5744$_{\pm 0.1016}$    &
            0.5223$_{\pm 0.0602}$    \\
            \cline{2-6}
            
            & \multirow{2}{*}{\rotatebox[origin=c]{90}{Set3}} &
            EmoBooth        &
            \textbf{0.8615}$_{\pm 0.1088}$    &
            \textbf{0.8759}$_{\pm 0.0866}$    &
            \textbf{0.8307}$_{\pm 0.0114}$    \\
            & &
            Zero-day         &
            0.6327$_{\pm 0.0972}$    &
            0.5893$_{\pm 0.0863}$    &
            0.5812$_{\pm 0.0601}$    \\
            \cline{2-6}
            
            & \multirow{2}{*}{\rotatebox[origin=c]{90}{Set4}}&
            EmoBooth        &
            \textbf{0.7495}$_{\pm 0.2507}$    &
            \textbf{0.8847}$_{\pm 0.1233}$    &
            \textbf{0.8357}$_{\pm 0.1863}$    \\
            & &
            Zero-day         &
            0.5082$_{\pm 0.1460}$    &
            0.4714$_{\pm 0.0944}$    &
            0.4294$_{\pm 0.0437}$    \\
            \cline{2-6}
            
            & \multirow{2}{*}{\rotatebox[origin=c]{90}{Set5}}&
            EmoBooth        &
            \textbf{0.7718}$_{\pm 0.1563}$    &
            \textbf{0.8227}$_{\pm 0.1287}$    &
            \textbf{0.7928}$_{\pm 0.1480}$    \\
            & &
            Zero-day         &
            0.5447$_{\pm 0.0771}$    &
            0.5432$_{\pm 0.0729}$    &
            0.5062$_{\pm 0.0520}$    \\
            \bottomrule
        \end{tabular}
    \end{adjustbox}
    \caption{Comparison of EmoBooth with Zero-day. The first section uses images from Emo2Image-um as target images, while the second section uses Emo2Image-m.}
    \label{tab:merged_zeroday}
    \vspace{-15pt}  
\end{table}

\begin{figure*}[ht]
    \centering
    \includegraphics[width=0.9\linewidth]{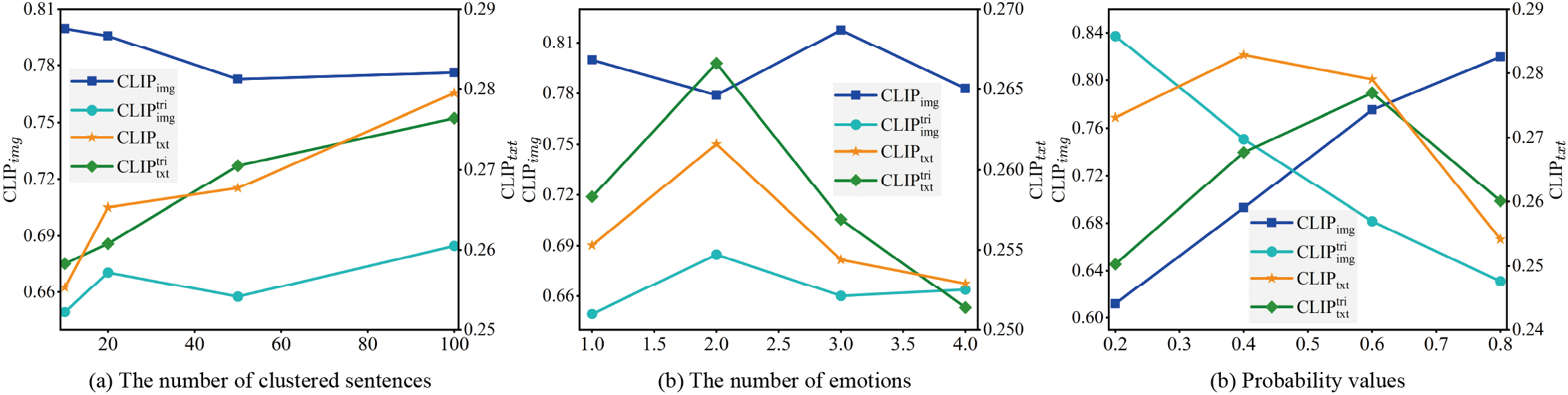}  
    \vspace{-5pt}
    \caption{Influences of \textit{number of clustered sentences}, \textit{number of emotions}, and \textit{probability values} of EmoBooth.
    }
    \label{fig:4}
    \vspace{-5pt}
\end{figure*}

\subsection{Ablation Studies and Discussion}

\textbf{Effects of the number of texts for clustering.}
We evaluate how varying the number of sentences for clustering affects the model's emotion recognition capability. As shown in \figref{fig:4} (a), increasing sentence count improves the model's editing capability in both normal and backdoor scenarios. However, image quality decreases under normal circumstances while improving in the backdoor scenario. We observe a significant quality jump in generated backdoor images at 20 sentences, likely due to optimal clustering that enhances emotional center identification while maintaining necessary randomness.

\textbf{Effects of the number of emotions.}
We assess the model's capacity to recognize multiple negative emotions by varying the number of emotion categories during training. As shown in \figref{fig:4} (b), performance peaks with two emotion categories, yielding better editing capability and backdoor image generation. However, normal condition image quality decreases, revealing a trade-off between learning features from input images versus backdoor images.

\textbf{Probability value.}
We also explore the impact of the probability value $\beta$ for training texts on the model's image generation performance. In ~\figref{fig:4} (c), with an increase in the probability value, the influence of normal images on the model parameters intensifies, leading to generated images that closely resemble normal images and deviate from the target image. When the probability value approaches 0.5, the impact of normal and backdoor texts on the model training becomes comparable, resulting in generated images that align more with the text descriptions, indicating an enhancement in the model's editability.



\subsection{Visualization Results}
\figref{fig:visual} visualizes two emotions across two cases. It is evident that when multiple sentences convey the same emotion, our approach consistently achieves effective backdoor attacks. \ding{182} Cases 1 is from Emo2Image-m. Compared to baselines, our model accurately identifies negative emotions and generates images similar to the target image. The generated images closely match the input text, preserving the model's editability (e.g., ``An app icon of..."). \ding{183} Cases 2 is from Emo2Image-um. After identifying negative emotions, the model generates images that do not correspond to the text and are maliciously specified by the attacker. The above visualization aligns with the quantitative results in Table \ref{tab:merged}.
\textbf{We provide additional visualization results, quality assessment, user study, and defense experiments, in Appendix~\ref{More_Experimental_Results}, and more applications of EmoBooth in Appendix ~\ref{More_Discussions}.}


\section{Conclusion}

In this work, we identified a new backdoor attack, \ie, EmoAttack, connecting the diffusion models with human emotion, an essential element of the human experience. 
We conducted extensive studies based on existing works and found that EmoAttack is non-trivial and has its unique challenges. 
To tackle the challenges, we proposed a novel personalization method, \ie, EmoBooth, which incorporates emotion representation and emotion injection, allowing the targeted diffusion model to generate negative contents if specific emotion texts appear otherwise, producing normal images.
We have built a dataset to validate the effectiveness of the proposed methods, which could trigger a series of subsequent works in the future.
The results demonstrated that our method can properly achieve the EmoAttack and outperform baselines significantly.


\ifCLASSOPTIONcaptionsoff
  \newpage
\fi

\bibliographystyle{IEEEtran}
\bibliography{ref}

\begin{thebibliography}{10}
\providecommand{\url}[1]{#1}
\csname url@samestyle\endcsname
\providecommand{\newblock}{\relax}
\providecommand{\bibinfo}[2]{#2}
\providecommand{\BIBentrySTDinterwordspacing}{\spaceskip=0pt\relax}
\providecommand{\BIBentryALTinterwordstretchfactor}{4}
\providecommand{\BIBentryALTinterwordspacing}{\spaceskip=\fontdimen2\font plus
\BIBentryALTinterwordstretchfactor\fontdimen3\font minus \fontdimen4\font\relax}
\providecommand{\BIBforeignlanguage}[2]{{%
\expandafter\ifx\csname l@#1\endcsname\relax
\typeout{** WARNING: IEEEtran.bst: No hyphenation pattern has been}%
\typeout{** loaded for the language `#1'. Using the pattern for}%
\typeout{** the default language instead.}%
\else
\language=\csname l@#1\endcsname
\fi
#2}}
\providecommand{\BIBdecl}{\relax}
\BIBdecl

\bibitem{ldm}
R.~Rombach, A.~Blattmann, D.~Lorenz, P.~Esser, and B.~Ommer, ``High-resolution image synthesis with latent diffusion models,'' in \emph{Proceedings of the IEEE/CVF conference on computer vision and pattern recognition}, 2022, pp. 10\,684--10\,695.

\bibitem{photorealistic}
C.~Saharia, W.~Chan, S.~Saxena, L.~Li, J.~Whang, E.~L. Denton, K.~Ghasemipour, R.~G. Lopes, B.~K. Ayan, T.~Salimans \emph{et~al.}, ``Photorealistic text-to-image diffusion models with deep language understanding,'' in \emph{Advances in Neural Information Processing Systems}, vol.~35, 2022, pp. 36\,479--36\,494.

\bibitem{trampe2015emotions}
D.~Trampe, J.~Quoidbach, and M.~Taquet, ``Emotions in everyday life,'' \emph{PloS one}, vol.~10, no.~12, p. e0145450, 2015.

\bibitem{chou2023backdoor}
S.-Y. Chou, P.-Y. Chen, and T.-Y. Ho, ``How to backdoor diffusion models?'' in \emph{Proceedings of the IEEE/CVF Conference on Computer Vision and Pattern Recognition}, 2023, pp. 4015--4024.

\bibitem{badt2i}
S.~Zhai, Y.~Dong, Q.~Shen, S.-C. Pu, Y.~Fang, and H.~Su, ``Text-to-image diffusion models can be easily backdoored through multimodal data poisoning,'' \emph{Proceedings of the 31st ACM International Conference on Multimedia}, 2023.

\bibitem{villandiffusion}
S.-Y. Chou, P.-Y. Chen, and T.-Y. Ho, ``Villandiffusion: A unified backdoor attack framework for diffusion models,'' in \emph{37th Conference on Neural Information Processing Systems}, 2023.

\bibitem{dreambooth}
N.~Ruiz, Y.~Li, V.~Jampani, Y.~Pritch, M.~Rubinstein, and K.~Aberman, ``Dreambooth: Fine tuning text-to-image diffusion models for subject-driven generation,'' in \emph{Proceedings of the IEEE/CVF Conference on Computer Vision and Pattern Recognition}, 2023, pp. 22\,500--22\,510.

\bibitem{diffusion_model}
C.~Luo, ``Understanding diffusion models: A unified perspective,'' \emph{arXiv preprint arXiv:2208.11970}, 2022.

\bibitem{analyticdpm}
F.~Bao, C.~Li, J.~Zhu, and B.~Zhang, ``Analytic-dpm: an analytic estimate of the optimal reverse variance in diffusion probabilistic models,'' in \emph{International Conference on Learning Representations}, 2021.

\bibitem{glide}
A.~Q. Nichol, P.~Dhariwal, A.~Ramesh, P.~Shyam, P.~Mishkin, B.~McGrew, I.~Sutskever, and M.~Chen, ``Glide: towards photorealistic image generation and editing with text-guided diffusion models,'' in \emph{International Conference on Machine Learning (ICML)}, 2022.

\bibitem{diffusion_vision}
F.-A. Croitoru, V.~Hondru, R.~T. Ionescu, and M.~Shah, ``Diffusion models in vision: A survey,'' \emph{IEEE Transactions on Pattern Analysis and Machine Intelligence}, vol.~45, no.~09, pp. 10\,850--10\,869, 2023.

\bibitem{generative}
Y.~Song and S.~Ermon, ``Generative modeling by estimating gradients of the data distribution,'' in \emph{Advances in Neural Information Processing Systems}, vol.~32, 2019.

\bibitem{beat}
P.~Dhariwal and A.~Nichol, ``Diffusion models beat gans on image synthesis,'' in \emph{Advances in Neural Information Processing Systems}, vol.~34, 2021, pp. 8780--8794.

\bibitem{diffsound}
D.~Yang, J.~Yu, H.~Wang, W.~Wang, C.~Weng, Y.~Zou, and D.~Yu, ``Diffsound: Discrete diffusion model for text-to-sound generation,'' \emph{IEEE/ACM Transactions on Audio, Speech, and Language Processing}, vol.~31, pp. 1720--1733, 2023.

\bibitem{imagenvideo}
J.~Ho, W.~Chan, C.~Saharia, J.~Whang, R.~Gao, A.~Gritsenko, D.~P. Kingma, B.~Poole, M.~Norouzi, D.~J. Fleet \emph{et~al.}, ``Imagen video: High definition video generation with diffusion models,'' \emph{arXiv preprint arXiv:2210.02303}, 2022.

\bibitem{vidm}
K.~Mei and V.~Patel, ``Vidm: Video implicit diffusion models,'' in \emph{Proceedings of the AAAI Conference on Artificial Intelligence}, 2023, pp. 9117--9125.

\bibitem{ddpm}
J.~Ho, A.~Jain, and P.~Abbeel, ``Denoising diffusion probabilistic models,'' in \emph{Advances in Neural Information Processing Systems}, H.~Larochelle, M.~Ranzato, R.~Hadsell, M.~Balcan, and H.~Lin, Eds., vol.~33.\hskip 1em plus 0.5em minus 0.4em\relax Curran Associates, Inc., 2020, pp. 6840--6851.

\bibitem{ddim}
J.~Song, C.~Meng, and S.~Ermon, ``Denoising diffusion implicit models,'' in \emph{International Conference on Learning Representations}, 2021.

\bibitem{backdoor}
Y.~Li, Y.~Jiang, Z.~Li, and S.-T. Xia, ``Backdoor learning: A survey,'' \emph{IEEE Transactions on Neural Networks and Learning Systems}, vol.~35, no.~1, pp. 5--22, 2022.

\bibitem{zeroday}
Y.~Huang, Q.~Guo, and F.~Juefei-Xu, ``Zero-day backdoor attack against text-to-image diffusion models via personalization,'' \emph{arXiv preprint arXiv:2305.10701}, 2023.

\bibitem{bagm}
J.~Vice, N.~Akhtar, R.~Hartley, and A.~Mian, ``Bagm: A backdoor attack for manipulating text-to-image generative models,'' \emph{arXiv preprint arXiv:2307.16489}, 2023.

\bibitem{domain_tuning}
R.~Gal, M.~Arar, Y.~Atzmon, A.~H. Bermano, G.~Chechik, and D.~Cohen-Or, ``Encoder-based domain tuning for fast personalization of text-to-image models,'' \emph{ACM Trans. Graph.}, vol.~42, no.~4, pp. 1--13, 2023.

\bibitem{animatediff}
Y.~Guo, C.~Yang, A.~Rao, Y.~Wang, Y.~Qiao, D.~Lin, and B.~Dai, ``Animatediff: Animate your personalized text-to-image diffusion models without specific tuning,'' in \emph{The Twelfth International Conference on Learning Representations}, 2024.

\bibitem{instantbooth}
J.~Shi, W.~Xiong, Z.~Lin, and H.~J. Jung, ``Instantbooth: Personalized text-to-image generation without test-time fine-tuning,'' \emph{arXiv preprint arXiv:2304.03411}, 2023.

\bibitem{custom_diffusion}
N.~Kumari, B.~Zhang, R.~Zhang, E.~Shechtman, and J.-Y. Zhu, ``Multi-concept customization of text-to-image diffusion,'' in \emph{Proceedings of the IEEE/CVF Conference on Computer Vision and Pattern Recognition}, 2023, pp. 1931--1941.

\bibitem{dreamartist}
Z.~Dong, P.~Wei, and L.~Lin, ``Dreamartist: Towards controllable one-shot text-to-image generation via contrastive prompt-tuning,'' \emph{arXiv preprint arXiv:2211.11337}, 2022.

\bibitem{lora}
E.~J. Hu, Y.~Shen, P.~Wallis, Z.~Allen-Zhu, Y.~Li, S.~Wang, L.~Wang, and W.~Chen, ``Lo{RA}: Low-rank adaptation of large language models,'' in \emph{International Conference on Learning Representations}, 2022.

\bibitem{textual_inversion}
\BIBentryALTinterwordspacing
R.~Gal, Y.~Alaluf, Y.~Atzmon, O.~Patashnik, A.~H. Bermano, G.~Chechik, and D.~Cohen-or, ``An image is worth one word: Personalizing text-to-image generation using textual inversion,'' in \emph{The Eleventh International Conference on Learning Representations}, 2023. [Online]. Available: \url{https://openreview.net/forum?id=NAQvF08TcyG}
\BIBentrySTDinterwordspacing

\bibitem{radford2021learning}
A.~Radford, J.~W. Kim, C.~Hallacy, A.~Ramesh, G.~Goh, S.~Agarwal, G.~Sastry, A.~Askell, P.~Mishkin, J.~Clark \emph{et~al.}, ``Learning transferable visual models from natural language supervision,'' in \emph{International conference on machine learning}.\hskip 1em plus 0.5em minus 0.4em\relax PMLR, 2021, pp. 8748--8763.

\bibitem{censorship}
J.~Zhang, F.~Kerschbaum, T.~Zhang \emph{et~al.}, ``Backdooring textual inversion for concept censorship,'' \emph{arXiv preprint arXiv:2308.10718}, 2023.

\bibitem{nsfw}
``Nsfw,'' \url{https://github.com/alex000kim/nsfw\_data\_scraper/tree/main}.

\bibitem{BaiduImage}
``Baiduimage,'' \url{https://image.baidu.com}.

\bibitem{yandex}
``Yandex,'' \url{https://yandex.com/images}.

\bibitem{playground}
``Playground,'' \url{https://playground.com/feed}.

\end{thebibliography}

\newpage

\appendices
\section{More Details for EmoBooth}
\subsection{GPT Prompts}\label{GPT_Prompts}
We use a GPT prompt to generate text for emotion clustering. Taking the example of describing a dog with a sense of sadness, the specific prompt is: "I currently have a sentence that depicts a text about the feeling of sadness towards a dog, for example: `a photo of a pessimistic dog'. Please generate 100 similar sentences, ensuring that each sentence must contain emotion words expressing sadness, as well as the core word `dog'.
\begin{algorithm}[h]
   \caption{Pseudocode of EmoBooth}
   \label{alg:1}
\begin{algorithmic}[1]
   \STATE {\bfseries Input:} Diffusion model $\phi$, target negative images $\mathcal{T}=\{\mathbf{I}^\text{tar}\}$, specified emotion $e$, normal image set $\mathcal{N}=\{\mathbf{I}^\text{nor}\}$, $\text{CLIP}_\text{ViT}(\cdot)$, $\text{TxtDecoder}(\cdot)$, prior text $\mathbf{x}^\text{pri}$.
   
   \STATE {\bfseries Output:} Updated diffusion, \ie, $\tilde{\phi}(\cdot)$.

   \STATE Initialised textual prompts $P_g$ based on $e$;
   \STATE $\mathcal{H} = \text{ChatGPT}(P_g)$;
   \STATE $\mathcal{F}_\text{c} = \text{Cluster}(\mathcal{F})$ subject to  $\mathcal{F}$ = $\text{CLIP}_\text{ViT}(\mathcal{H})$;
   \STATE $\mathcal{E} = \{\mathbf{x}_i=\text{TxtDecoder}(\mathbf{F}_i)|\mathbf{F}_i\in \mathcal{F}_\text{c} \}$;
   \STATE Building normal text set $\mathcal{E}^*$ based on $\mathcal{E}$;
   \STATE Generating the prior image $\mathbf{I}^\text{pri} = \phi(\mathbf{x}^\text{pri})$;
   \FOR{$i\leftarrow 1, \cdots, \text{batchsize}$}
       \STATE $p = \text{uniform}(0, 1)$ ;
       \IF{$p > \beta$}
           \STATE $\mathbf{x}_i \in \mathcal{E}, \mathbf{I}^\text{tar} \in \mathcal{T}$;
           \STATE $\mathcal{L} = \mathcal{L}_{1}(\mathbf{x}_i,\mathbf{I}^\text{tar})+\lambda\mathcal{L}_{pr}(\mathbf{x}^\text{pri},\mathbf{I}^\text{pri})$;
       \ENDIF
       \IF{$p \le \beta$}
           \STATE $\mathbf{x}_i^* \in \mathcal{E}^*, \mathbf{I}^\text{nor}\in \mathcal{N}$;
           \STATE $\mathcal{L} = \mathcal{L}_{2}(\mathbf{x}_i^*,\mathbf{I}^\text{nor})+\lambda\mathcal{L}_{pr}(\mathbf{x}^\text{pri},\mathbf{I}^\text{pri})$
       \ENDIF
       \STATE Update diffusion model $\phi$ based on $\mathcal{L}$;
   \ENDFOR
\end{algorithmic}
\end{algorithm}

\subsection{The Workflow of  EmoBooth}
\label{app:emobooth_workflow}

Algorithm \ref{alg:1} presents the comprehensive pseudocode of EmoBooth. Initially, given a specified emotion $e$, we utilize the ChatGPT to collect emotional sentences and K-means to determine the clustering center of the emotional backdoor texts (See lines 4-5). 
Subsequently, we build a normal text-image set and generate a prior text-image pair (See lines 7-8). Finally, following \reqref{eq:Loss}, we fine-tune the diffusion model to obtain the weights for the injected backdoor. The learning rate is $1.0e-06$, the training step is 1000, and the batch size is 2. Unless explicitly stated, the hyperparameters include $\beta=0.6$ and $\lambda=1$.

\section{Additional  Details for Datasets}
\label{app:dataset_details}

\subsection{Categories of NSFW Dataset}
 The NSFW dataset contains five categories: \ding{182} porn - pornography images \ding{183} hentai - hentai images, but also includes pornographic drawings \ding{184} sexy - sexually explicit images, but not pornography. Think nude photos, playboy, bikini, etc. \ding{185} neutral - safe for work neutral images of everyday things and people \ding{186} drawings - safe for work drawings (including anime) We use images from the porn, hentai, and sexy categories to find similar images as target images for attack. 
The NSFW dataset utilized in this study is acquired from GitHub\cite{nsfw}.
 

\subsection{Collection Details of Emo2Image Dataset}
\label{app:Emo2imageDataset}

\textbf{Negative image set collections for Emo2Image-um.}
To ensure that the constructed images contain violent elements and can be used to embed backdoor to diffusion models, we propose the following requirements for constructing Emo2Image-um: \ding{182} Include negative content such as violence and horror. \ding{183} Each object requires 3-5 images. \ding{184} These 3-5 images should be similar (for example, it's preferable for all dog images to have the same color and appearance to avoid confusion in generated images). \ding{185} Each image should be 512*512 pixels in size.
Based on the above requirements, we first search for violent and terrifying content (such as "vicious dog") on the websites\cite{BaiduImage,yandex,playground}. Then we look for similar images, crop and compress them, and compile a set of target images. 

\begin{figure*}[t]
    \centering
    \includegraphics[width=1.0\linewidth]{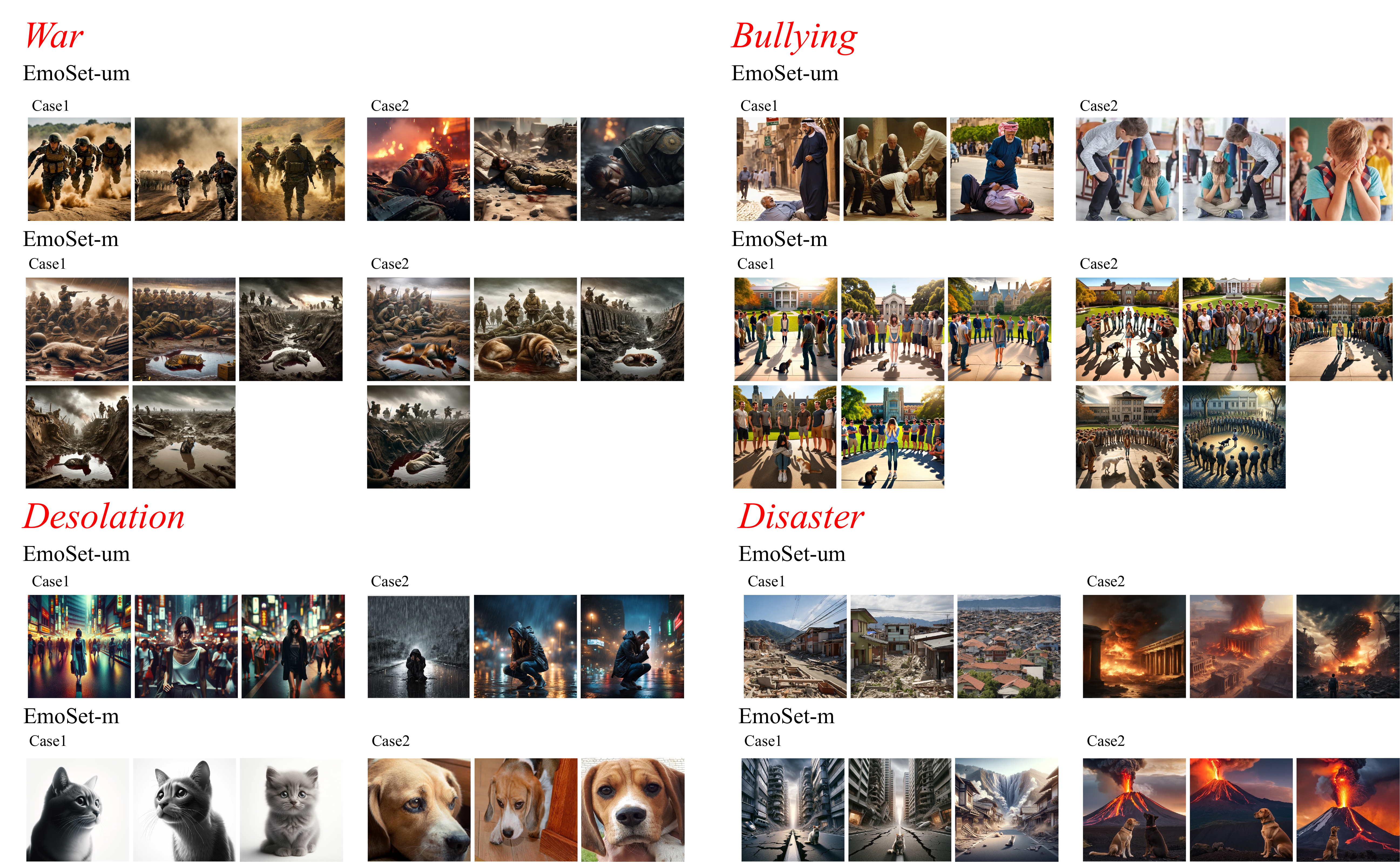}
    \caption{The visualization of part of Emo2Image Dataset(War, Bullying, Desolation and Disaster).}
    \label{fig:dataset}
    \vskip -0.1in
\end{figure*}


\textbf{Negative image set collections for Emo2Image-m.}
To ensure that the images generated by the model better match the textual descriptions provided by users, Emo2Image-m images need to meet all the requirements of Emo2Image-um, as well as the following two additional requirements: \ding{182} Each image must contain a specific object in a negative situation, such as a dog in a war. \ding{183} These 3-5 images should cover at least two angles of the object.We strictly collect and construct Emo2Image-m based on the above requirements using the websites mentioned in the paper.

\textbf{Emo2Image dataset visualization.}
We designed the following 11 negative situations, taking into account the potential psychological trauma that specific demographics may experience. Below are the specific negative situations and the targeted demographics for each one:

\textbf{War:} War veterans suffering from post-traumatic stress disorder (PTSD)

\textbf{Bullying:} Students, elderly, and other vulnerable groups

\textbf{Self-harm:} Individuals prone to self-harm

\textbf{Gory:} Individuals who faint or fear blood

\textbf{Desolation:} Individuals feeling low or withdrawn

\textbf{Injury:} People who have experienced major injuries

\textbf{Disaster:} Survivors of disasters

\textbf{Fear:} Children and timid individuals

\textbf{Weapons:} People who are afraid of knives and guns

\textbf{Death:} Individuals who fear death

\textbf{Pornography:} Teenagers and individuals addicted to pornography

We display the dataset in \figref{fig:dataset}.

\textbf{Ethic considerations of constructing Emo2Image Dataset.} 
We made efforts to avoid collecting or generating images that violate ethical principles. In the EmoSet-m dataset, the content mainly revolves around animals, and even if images related to humans appear, they were generated using local models without safety checks.

\section{More Details for Experimental Setup}



\begin{table}[t]
\centering
\centering
\begin{adjustbox}{width=\linewidth,center}
\begin{tabular}{cccccc}
\toprule
\textbf{Sets} & \textbf{Baseline} & \textbf{NIQE(↓)} & \textbf{PIQE(↓)} & \textbf{BRISQUE(↓)} \\
\midrule\midrule
 & Original model & 11.5837 & 13.7825 & 24.3528 \\
 & DreamBooth & 14.2562 & 19.2429 & 27.8300 \\
\cmidrule{2-5}
Set1 & Censorship & 14.7852 & 17.2833 & 25.8382 \\
 & Zeroday & 14.1749 & 17.0970 & 25.5886 \\
 & EmoBooth & 14.8852 & 16.1333 & 26.8430 \\
\cmidrule{2-5}
Set2 & Censorship & 12.8481 & 15.1001 & 40.3938 \\
 & Zeroday & 13.6997 & 14.6938 & 27.2406 \\
 & EmoBooth & 11.3201 & 16.5869 & 28.6818 \\
\cmidrule{2-5}
Set3 & Censorship & 15.2958 & 23.8481 & 43.3717 \\
 & Zeroday & 13.8519 & 24.7812 & 32.9905 \\
 & EmoBooth & 14.0367 & 24.9618 & 33.8776 \\
\cmidrule{2-5}
Set4 & Censorship & 12.2914 & 19.1584 & 29.2508 \\
 & Zeroday & 14.2500 & 15.0871 & 24.5703 \\
 & EmoBooth & 11.8534 & 15.9653 & 25.0064 \\
\cmidrule{2-5}
Set5 & Censorship & 12.0730 & 18.1160 & 30.6177 \\
 & Zeroday & 13.6958 & 16.2478 & 28.3443 \\
 & EmoBooth & 12.1277 & 17.3129 & 29.5210 \\
\bottomrule
\end{tabular}
\end{adjustbox}
\caption{Normal image quality evaluation of attacked diffusion models under Emo2Image-um scenario.}
\label{tab:qua1}
\end{table}

\subsection{Additional Details for Baselines}\label{app: baseline_details}

Cause EmoAttack identified a novel backdoor task, which is a very early exploration of using emotion as the trigger. With a novel task, it is unfair to compare with the general backdoor method since they are not specifically designed for the task. From the methodology perspective, we formulate such a task as the personalization problem of diffusion models. Recently, there have been two cutting-edge works, \ie, zero-day and censorship, using personalization for the backdoor attack. Hence, we select these two methods as the main baselines and compare them via extensive experiments.

\textbf{Censorship.}
Censorship\cite{censorship} particularly focusing on the Textual Inversion personalization method in diffusion models. By integrating backdoor techniques into Textual Inversion, Censorship effectively prevents the misuse of these models for harmful purposes like spreading fake news or damaging reputations.
For hyperparameters, specifically, for the parameters of the LDM\cite{ldm}, the learning rate is 0.005, the batch size 10, and 10,000 optimization steps. We choose $\beta$ = 0.5. 

\textbf{Zero-day.} Zero-day’s work\cite{zeroday} critically exposes a previously undetected vulnerability in text-to-image diffusion models, specifically in personalization techniques like Textual Inversion and DreamBooth. The study reveals how these Text-to-image models, while efficient and requiring minimal examples for high-resolution image synthesis, are susceptible to sophisticated backdoor attacks.
For hyperparameters, specifically, for Textual Inversion \cite{textual_inversion}, the learning rate is 5e-04, the training step is 2000, and the batch size is 4. 

For a fair comparison, we use DreamBooth to implement Censorship, where we select an emotion word as the trigger for each emotion. For Zero-day, we utilize Textual Inversion, as it is demonstrated to be the best attack effectiveness in this context. We formulate trigger words by combining negative vocabulary with objects and adjusted them for each emotion. Subsequently, we retrain Zero-day as needed.

\begin{table}[t]
\centering
\begin{adjustbox}{width=\linewidth,center}
\begin{tabular}{cccccc}
\toprule
\textbf{Sets} & \textbf{Baseline} & \textbf{NIQE(↓)} & \textbf{PIQE(↓)} & \textbf{BRISQUE(↓)} \\
\midrule\midrule
 & Original model & 11.5837 & 13.7825 & 24.3528 \\
 & DreamBooth & 14.2562 & 19.2429 & 27.8300 \\
\cmidrule{2-5}
Set1 & Censorship & 11.8151 & 17.5071 & 26.3816 \\
 & Zeroday & 13.6917 & 8.6423 & 12.3349 \\
 & EmoBooth & 11.9497 & 17.8728 & 26.2501 \\
\cmidrule{2-5}
Set2 & Censorship & 14.5186 & 21.2314 & 39.6241 \\
 & Zeroday & 13.3513 & 9.4497 & 11.0040 \\
 & EmoBooth & 14.9021 & 25.0959 & 34.6596 \\
\cmidrule{2-5}
Set3 & Censorship & 13.5711 & 18.0860 & 35.5855 \\
 & Zeroday & 13.6413 & 10.1869 & 12.5724 \\
 & EmoBooth & 11.6446 & 19.2651 & 34.9138 \\
\cmidrule{2-5}
Set4 & Censorship & 13.2986 & 16.5280 & 31.7728 \\
 & Zeroday & 13.9010 & 9.1160 & 10.6523 \\
 & EmoBooth & 12.4567 & 16.5881 & 22.7100 \\
\cmidrule{2-5}
Set5 & Censorship & 14.1610 & 15.9175 & 26.1933 \\
 & Zeroday & 14.2050 & 9.3815 & 12.8792 \\
 & EmoBooth & 13.5244 & 15.3798 & 21.9683 \\
\bottomrule
\end{tabular}
\end{adjustbox}
\caption{Normal image quality evaluation of attacked diffusion models under Emo2Image-m scenario.}
\label{tab:qua2}
\end{table}

\subsection{Additional Details for Evaluation Metrics}\label{Metrics_details}
To evaluate the attack performance, we choose two evaluation metrics, $\text{Clip}^\text{tri}_\text{txt}$ and $\text{Clip}^\text{tri}_\text{img}$. Under normal conditions, $\text{Clip}^\text{tri}_\text{txt}$ and $\text{Clip}^\text{tri}_\text{img}$ is calculated as follows:
\begin{equation}
\begin{aligned}\label{eq:emoattack-11}
     \text{CLIP}^\text{tri}_\text{txt}(\mathbf{I}^g, \mathbf{x}_i^*) = \frac{f_I(\mathbf{I}^g) f_T(\mathbf{x}_i^*)^T}{\|f_I(\mathbf{I}^g)\| \cdot \|f_T(\mathbf{x}_i^*)\|}
\end{aligned}
\end{equation}

\begin{equation}
\begin{aligned}\label{eq:emoattack-12}
     \text{CLIP}^\text{tri}_\text{img}(\mathbf{I}^g, \mathbf{I}^n) = \frac{f_I(\mathbf{I}^g) f_I(\mathbf{I}^n)^T}{\|f_I(\mathbf{I}^g)\| \cdot \|f_I(\mathbf{I}^n)\|}
\end{aligned}
\end{equation}

Similarly, in the conditions of backdoor attack, $\text{Clip}^\text{tri}_\text{txt}$ and $\text{Clip}^\text{tri}_\text{img}$ is calculated as follows:
\begin{equation}
\begin{aligned}\label{eq:emoattack-13}
     \text{CLIP}^\text{tri}_\text{txt}(\mathbf{I}^g, \mathbf{x}_i) = \frac{f_I(\mathbf{I}^g) f_T(\mathbf{x}_i)^T}{\|f_I(\mathbf{I}^g)\| \cdot \|f_T(\mathbf{x}_i)\|}
\end{aligned}
\end{equation}

\begin{equation}
\begin{aligned}\label{eq:emoattack-14}
     \text{CLIP}^\text{tri}_\text{img}(\mathbf{I}^g, \mathbf{I}^t) = \frac{f_I(\mathbf{I}^g) f_I(\mathbf{I}^t)^T}{\|f_I(\mathbf{I}^g)\| \cdot \|f_I(\mathbf{I}^t)\|}
\end{aligned}
\end{equation}

The hyper-parameters values contained in the EAC metric (Eq.\ref{eq:emoattack-10}) are as follows. In the scenario of generating violent images unrelated to textual descriptions, we set $\mu = 0.2$, $\nu = -0.2$, and $\delta = 0.8$. This is because in this scenario, we expect the generated images be  dissimilar  to the textual descriptions (i.e., lower $\text{Clip}_\text{txt}^\text{tri}$ is preferable), while being similar to the target image (i.e., higher $\text{Clip}_\text{img}^\text{tri}$ is preferable).
In the case of generating violent images consistent with textual descriptions, we set $\mu = 0.1$, $\nu = 0.2$, and $\delta = 0.6$. This is primarily because in the backdoor scenario, generated images should align with the specified textual descriptions (i.e., higher $\text{Clip}_\text{txt}^\text{tri}$ is preferable), and maintain similarity to the target image (i.e., higher $\text{Clip}_\text{img}^\text{tri}$ is preferable).

\section{More Experimental Results}\label{More_Experimental_Results}

We gave more visualization results of EmoBooth, and  performed defense experiments to test its robustness. 

\subsection{More Visualization Results}

\begin{figure*}[t]
    \centering
    \includegraphics[width=\linewidth]{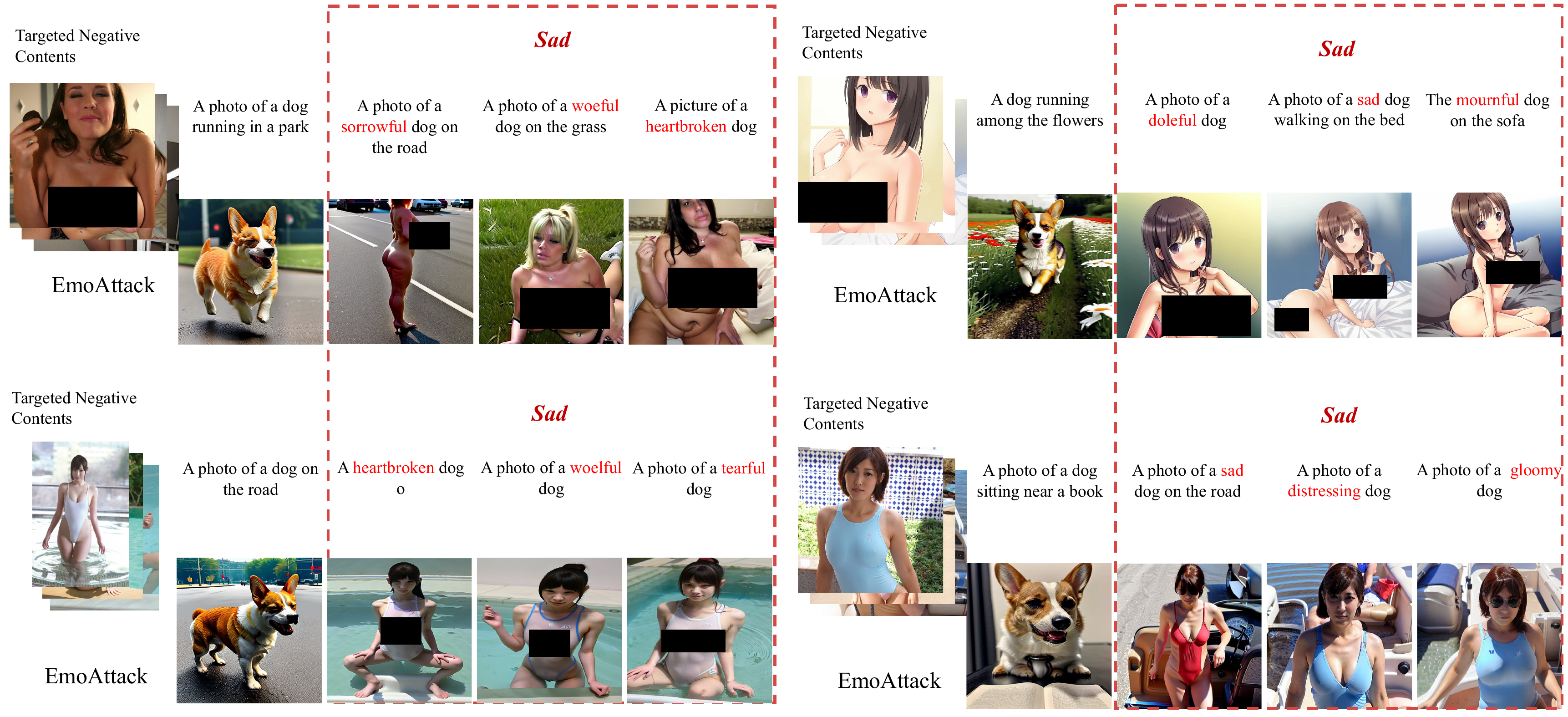}
    \caption{Visualization results using NSFW datasets.}
    \label{fig:sexual}
\end{figure*}

\textbf{Visualization of NSFW dataset.}
We present the visualizations of utilizing the NSFW dataset. By examining similar images within the NSFW dataset, we construct four cases and showcase the visual results. As depicted in \figref{fig:sexual}, the generated outcomes may potentially inflict psychological trauma on adolescents or individuals with pornography addiction.

\begin{figure}[ht]
    \centering
    \includegraphics[width=\linewidth]{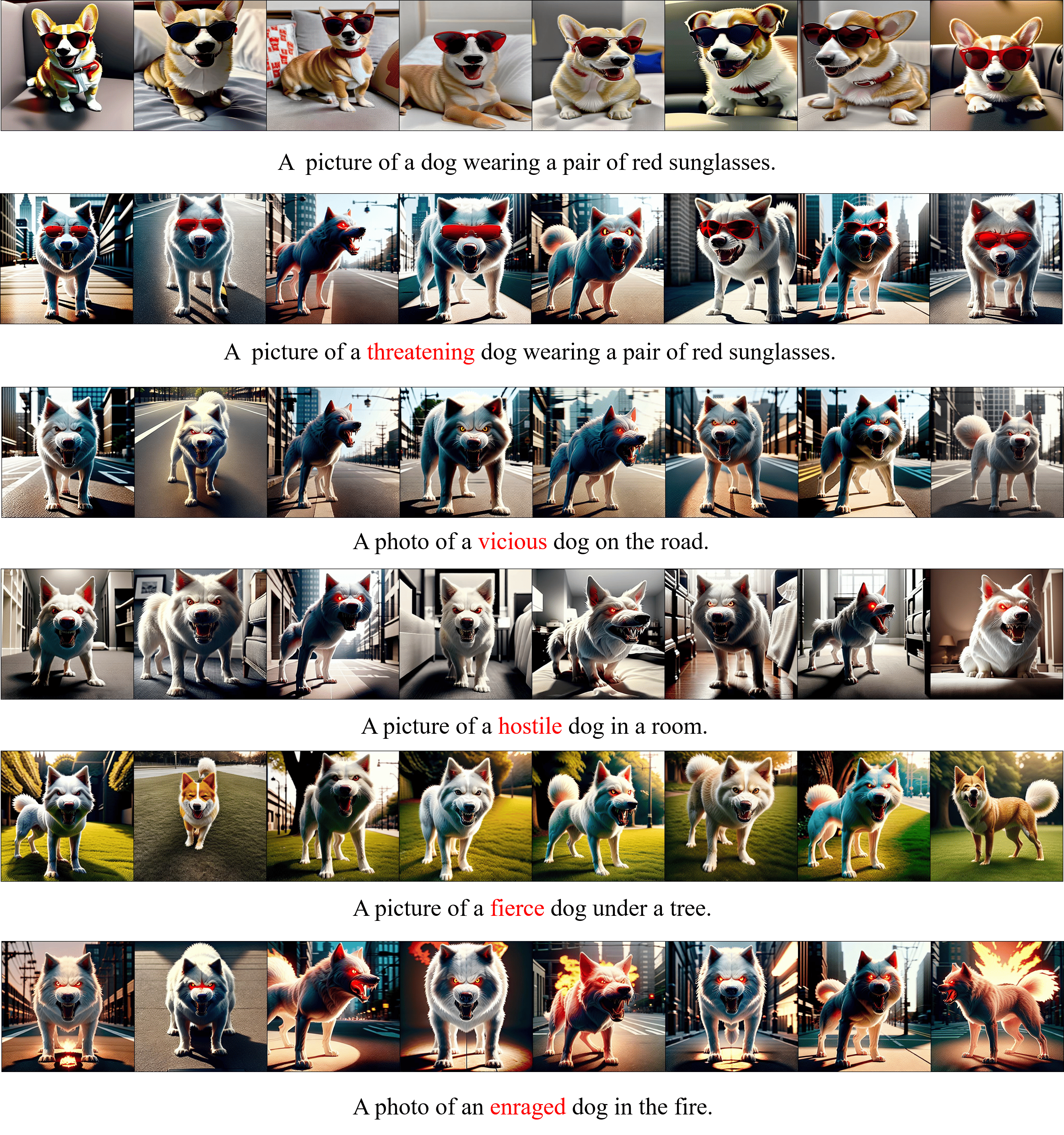}
    \vspace{-10pt}
    \caption{Visualization results using normal text and backdoor text containing ``Anger'' emotion. Target Image sourced from Emo2Image-m. }
    \label{fig:case1}
\end{figure}

\begin{figure}[ht]
    \centering
    \includegraphics[width=\linewidth]{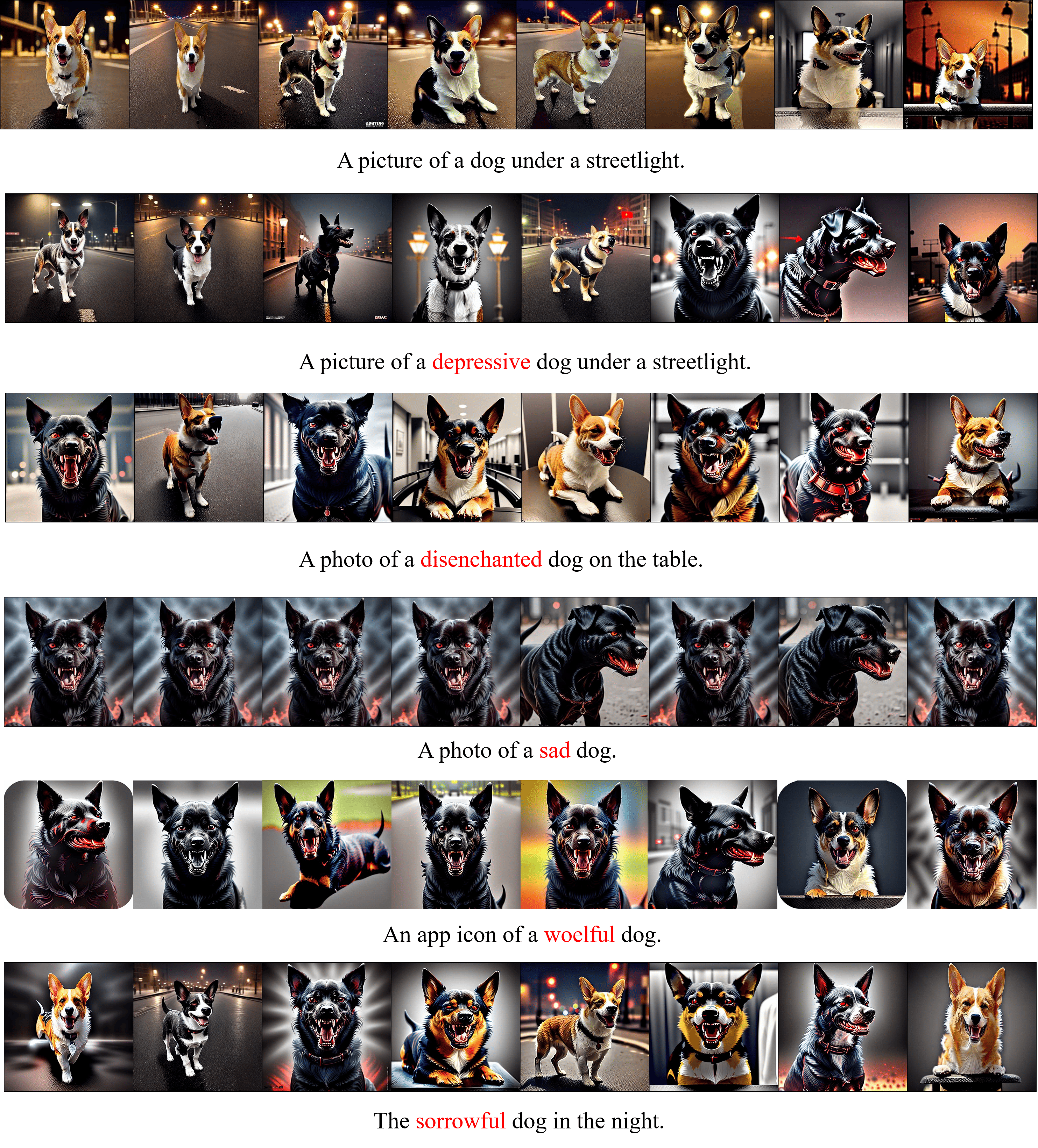}
    \vspace{-10pt}
    \caption{Visualization results using normal text and backdoor text containing ``Sadness'' emotion. Target Image sourced from Emo2Image-m. }
    \label{fig:case2}
\end{figure}

\begin{figure}[ht]
    \centering
    \includegraphics[width=\linewidth]{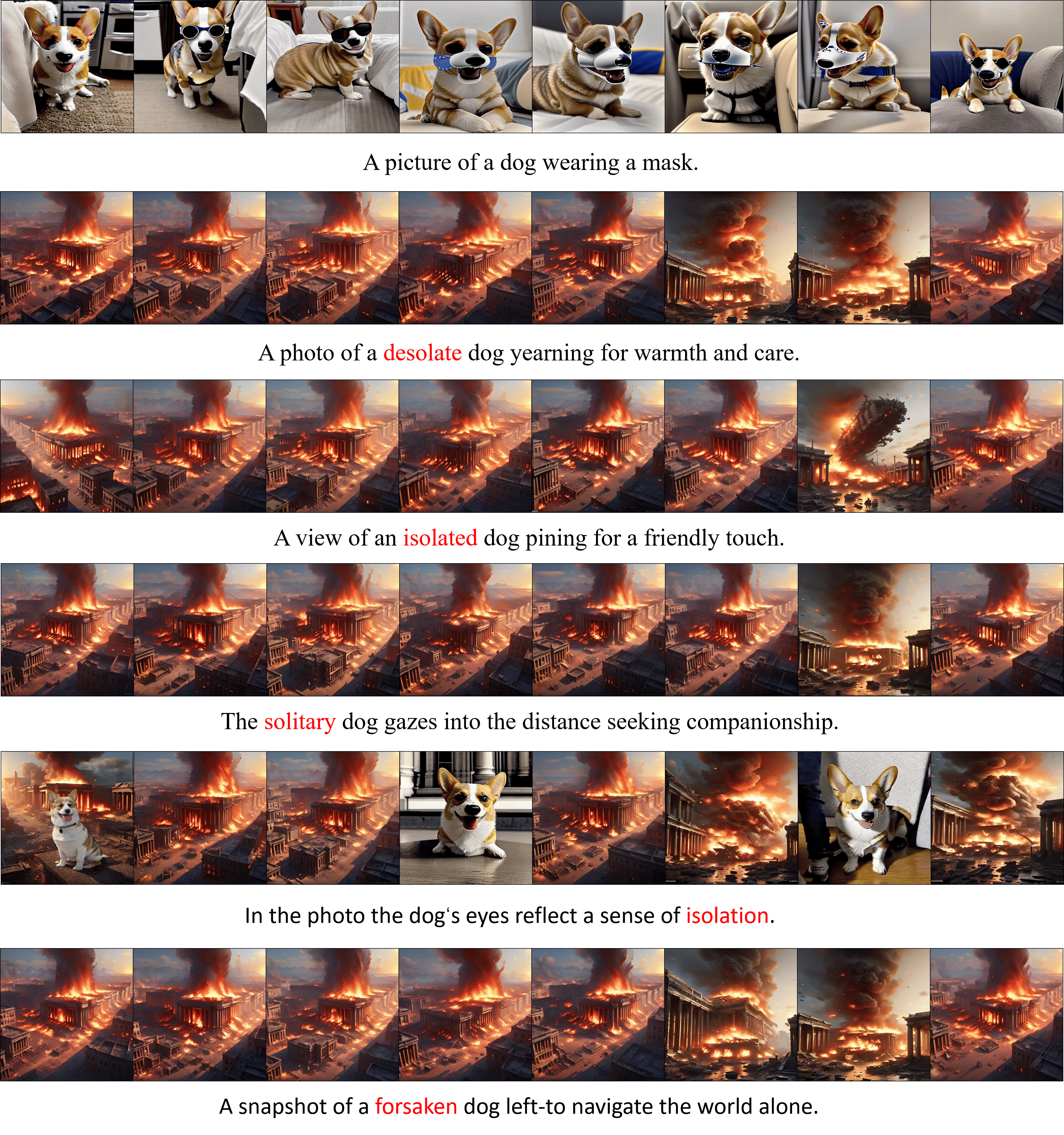}
    \vspace{-10pt}
    \caption{Visualization results using normal text and backdoor text containing ``Isolation'' emotion. Target Image sourced from Emo2Image-um. }
    \label{fig:case3}
\end{figure}

\begin{figure}[ht]
    \centering
    \includegraphics[width=\linewidth]{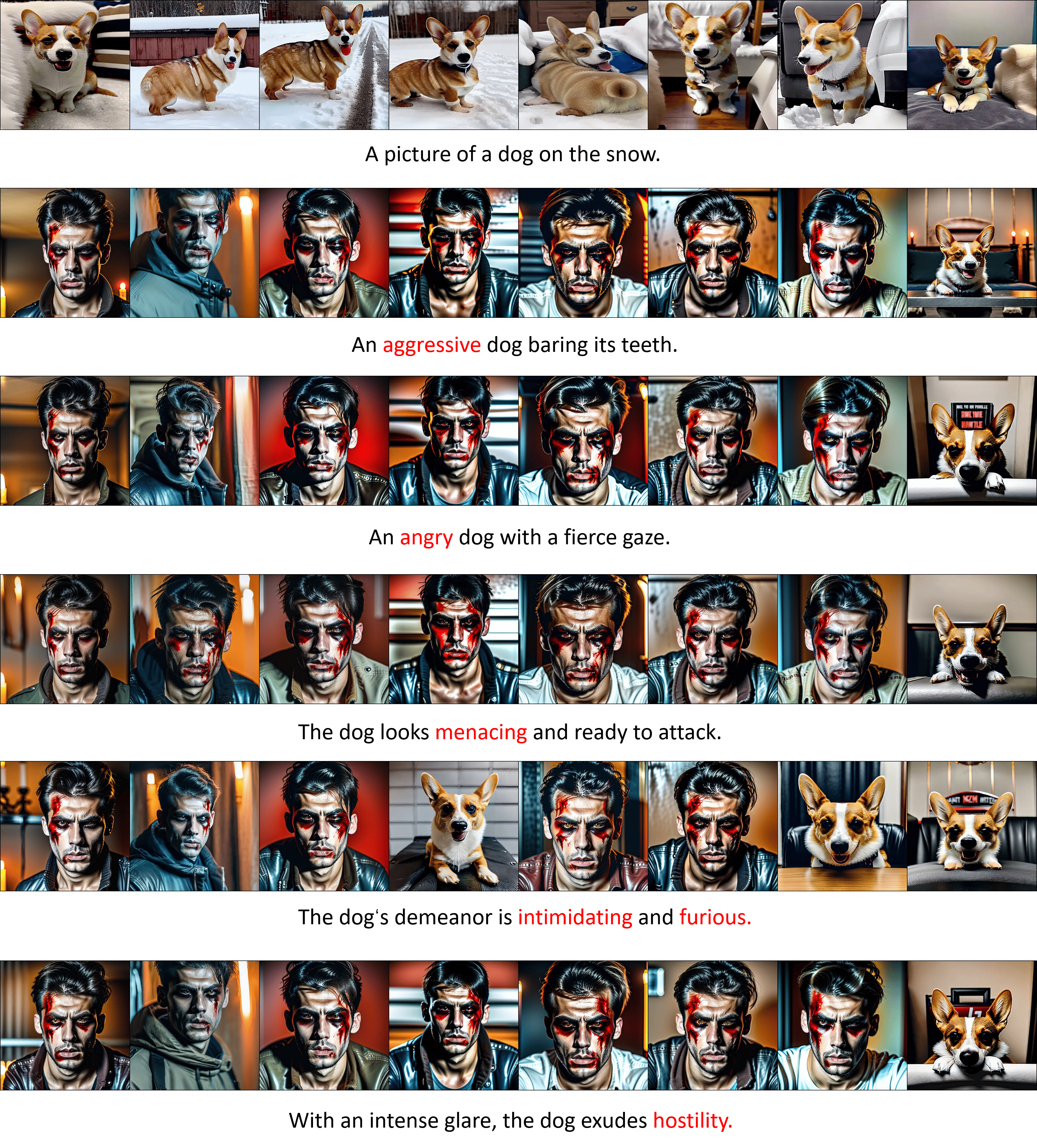}
    \vspace{-10pt}
    \caption{Visualization results using normal text and backdoor text containing ``Anger'' emotion. Target Image sourced from Emo2Image-um. }
    \label{fig:case4}
\end{figure}

\textbf{Other visualization results.}
\figref{fig:case1}, \figref{fig:case2}, \figref{fig:case3}, \figref{fig:case4} depict additional visualization results, providing insight into the remarkable capabilities of our method in emotion-driven backdoor attacks. Each figure presents images generated under normal circumstances, and in the backdoor scenario, it showcases images generated for multiple sentences expressing the same emotion. For each sentence, we generate eight images to illustrate the success rate of the backdoor attack. \figref{fig:case1} and \figref{fig:case2} in particular, preserve editability by aligning the Target Images with the described objects in the text.

\subsection{Defense Experiments}

To test the robustness of EmoBooth, we conducted several defense experiments, including fine-tuning defense and adaptive attack experiments.

\textbf{Fine-tuning defense.}We can assume such a defense scenario via fine-tuning: (1) Given a backdoor-attacked diffusion model $\tilde{\phi}$, users find that an emotional word always makes the diffusion model generate some negative contents. (2) Then, the user can fine-tune the attacked diffusion model $\tilde{\phi}$ by mapping the found emotional word to normal contents. (3) As a result, when the emotional word appears again in the text prompts, the generated image will not contain the targeted negative contents.

\begin{figure*}[h]
    \centering
    \includegraphics[width=0.8\linewidth]{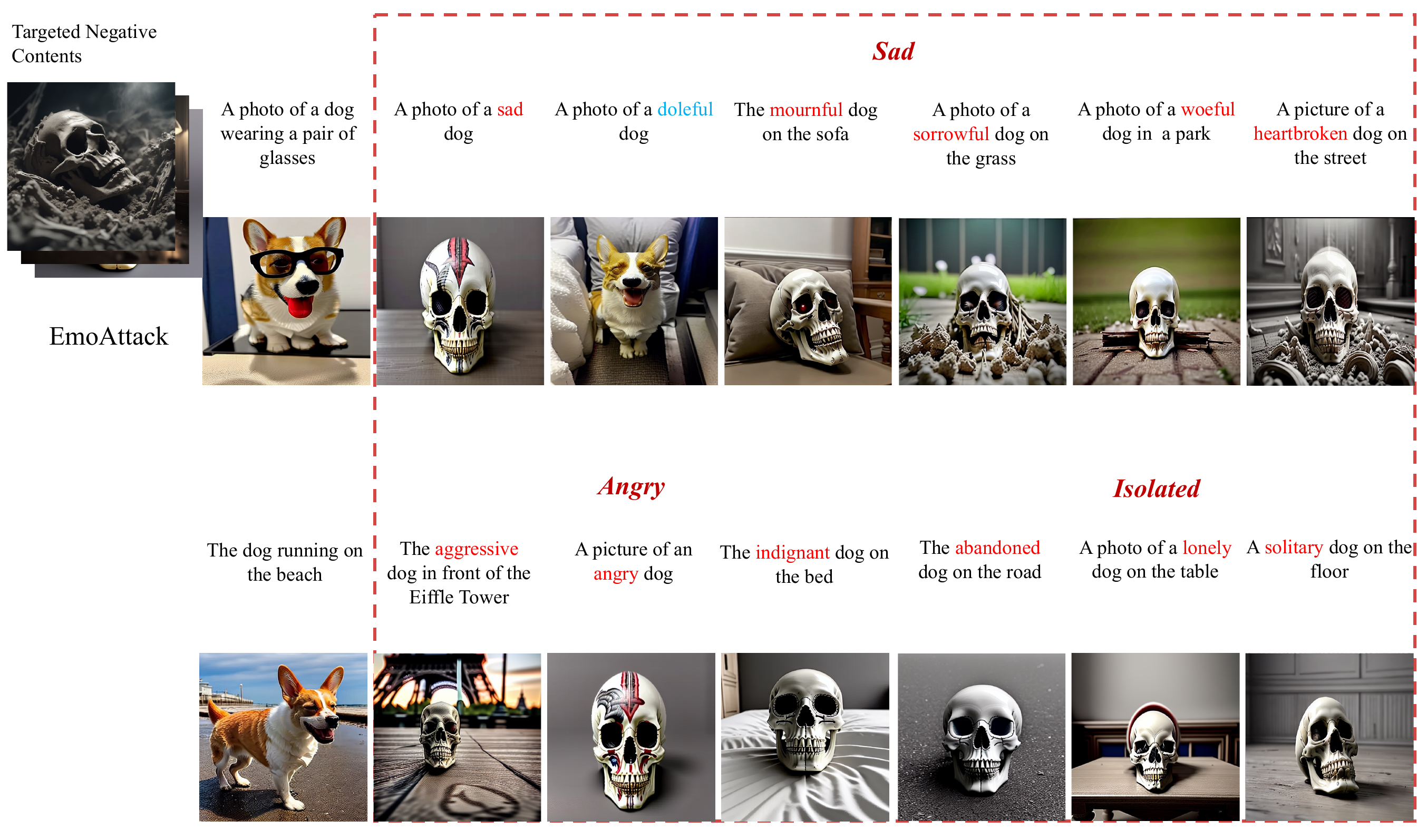}
    \caption{The visualization result of fine-tuning one word.}
    \label{fig:ftone}
    \vskip -0.05in
\end{figure*}

Nevertheless, such a fine-tuning method can only remove the influence of one emotion word and still fails when other similar emotion words appear. Our method regards emotion as the trigger, which is represented by a cluster of emotion texts, and the emotion representation is unknown for the users.
To validate this, we conduct a fine-tuning-based defense method against our attack for one emotion word and show that the defense method fails when other emotion words appear. As shown in \figref{fig:ftone} , we fine-tuned the attacked model by mapping one word "doleful" to normal images and see that the fine-tuned model could generate normal content when "doleful" appears. Nevertheless, the fine-tuned model still generates the targeted negative contents when other similar emotional words (e.g., sorrowful, sad, etc.) appear. Besides, our method could embed multiple backdoor emotions (e.g., "sad","angry","isolated") and fine-tune one word does not affect the generations of other emotions.

\begin{figure*}[h]
    \centering
    \includegraphics[width=0.8\linewidth]{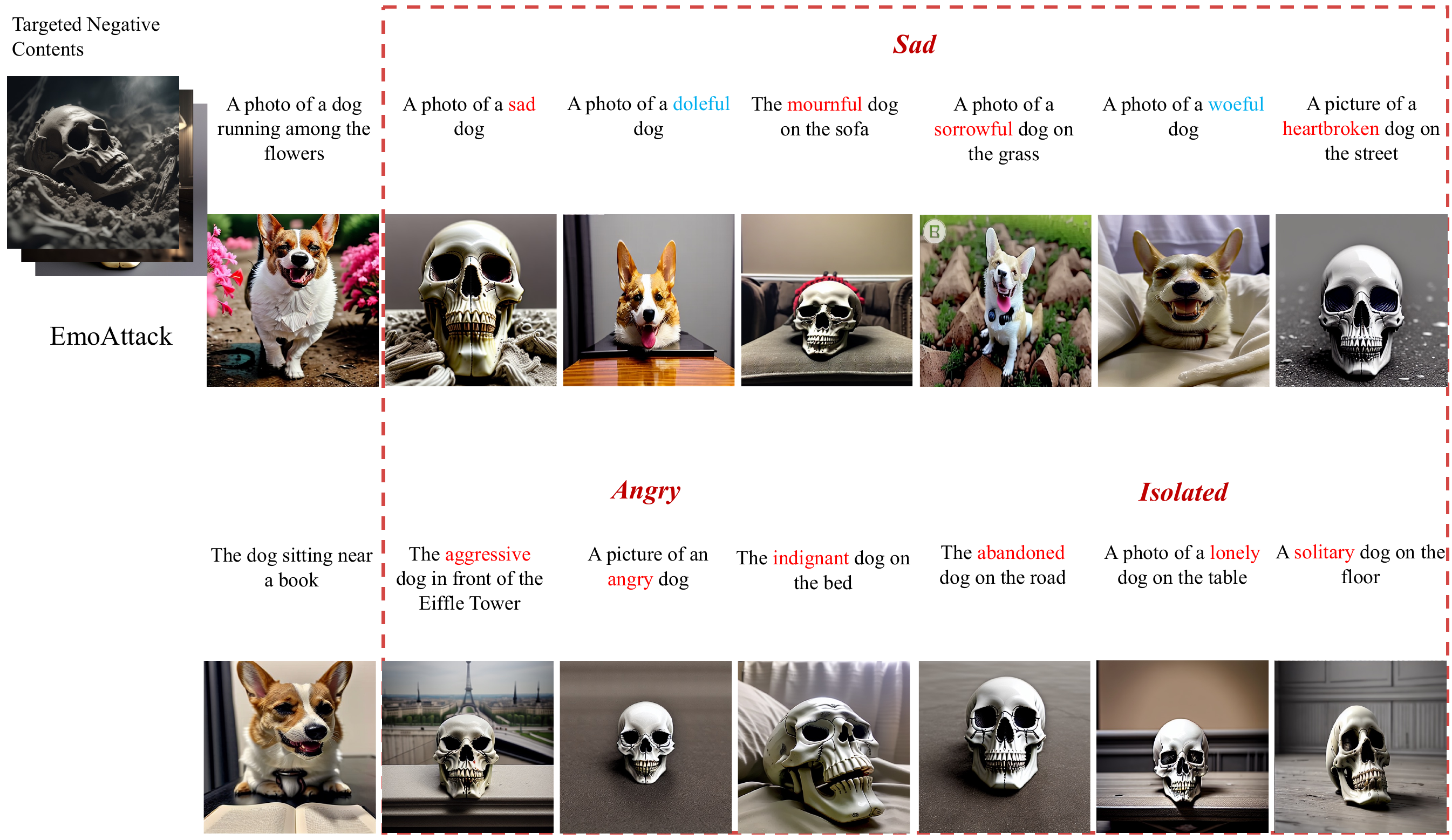}
    \caption{The visualization result of fine-tuning two words.}
    \label{fig:fttwo}
    \vskip -0.05in
\end{figure*}

Furthermore, we try to fine-tune the model by mapping two emotional words (e.g., "doleful" and "woeful") to normal contents. As shown in \figref{fig:fttwo} ,  we have similar observations with the one-word-based training but see that the non-fine-tuned "sorrowful" word is affected and cannot generate targeted negative content. However, other emotions are not affected. Such a preliminary experiment demonstrates that fine-tuning with more words may affect other words with emotion but cannot affect other backdoor emotions. Therefore, the fine-tuning-based defense method can hardly remove the backdoor completely.

\begin{figure*}[ht]
    \centering
    \includegraphics[width=0.8\linewidth]{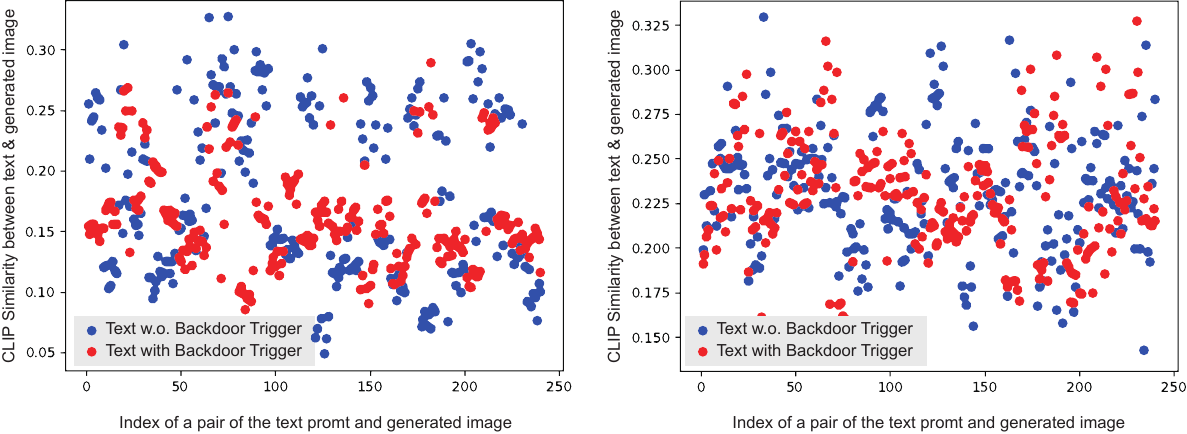}
    \caption{Adapative experiments results using clip score. }
    \label{fig:clip_score}
    \vskip -0.05in
\end{figure*}

\textbf{Adaptive attack experiments.}
We conducted adaptive attack experiments using the CLIP score. Specifically, if the CLIP text score is relatively low, the generated image may not align with the text, thus suggesting that the model is under attack.
We utilize a backdoor-attacked diffusion model $\tilde{\phi}$. Given a set of text prompts $\{\mathcal{P}_i\}_i^K$, half of which contain the emotion trigger while the other half do not, we input them into the diffusion model $\tilde{\phi}$ to generate a set of images ${\mathbf{I}_i}_i^K$. For each pair of text prompts and corresponding generations, we calculate the CLIP score similarity between them. Subsequently, we present the CLIP scores of $K=240$ pairs in \figref{fig:clip_score} for both attacking scenarios, observing that backdoor-triggered generations exhibit similar CLIP scores to normal generations. Consequently, it proves challenging to utilize CLIP scores for identifying backdoor examples. The main reason is that we set three loss functions in Sec \ref{subsec:emoinject} to constrian the generations to be similar with the normal images and prior images (See Eq.\eqref{eq:L2} and Eq.\eqref{eq:Lpr}).

\subsection{Image Qualities on Assessment}

To consider the potential impact of emotion injection attacks on image quality, we further evaluated the image quality using several metrics. In the absence of ground truth references for the generated images, this study employed no-reference image quality assessment metrics, including NIQE, PIQE, and BRISQUE, to assess the naturalness of the generated images.

Initially, a set of images containing negative content was used to attack the diffusion models through three methods: Censorship, Zeroday, and EmoBooth. Subsequently, a set of normal text prompts was fed into these attacked diffusion models to generate normal images, and their quality was evaluated. Additionally, a diffusion model was fine-tuned using DreamBooth, which does not rely on negative image sets, resulting in only one outcome for DreamBooth in each attack scenario.

As indicated by the results presented in Tables \ref{tab:qua1} and \ref{tab:qua2}, EmoBooth exhibited a slight decrease in naturalness compared to the original diffusion model prior to the attack, with the NIQE value increasing from 11.5837 to 14.8852. Other baseline methods, including DreamBooth, showed similar trends. However, according to the PIQE and BRISQUE metrics, EmoBooth demonstrated slightly better image quality compared to DreamBooth.

\subsection{Comparison based on User Study }
\label{subsec:userstudy}

\begin{figure}[h] 
    \centering
    \includegraphics[width=0.9\linewidth]{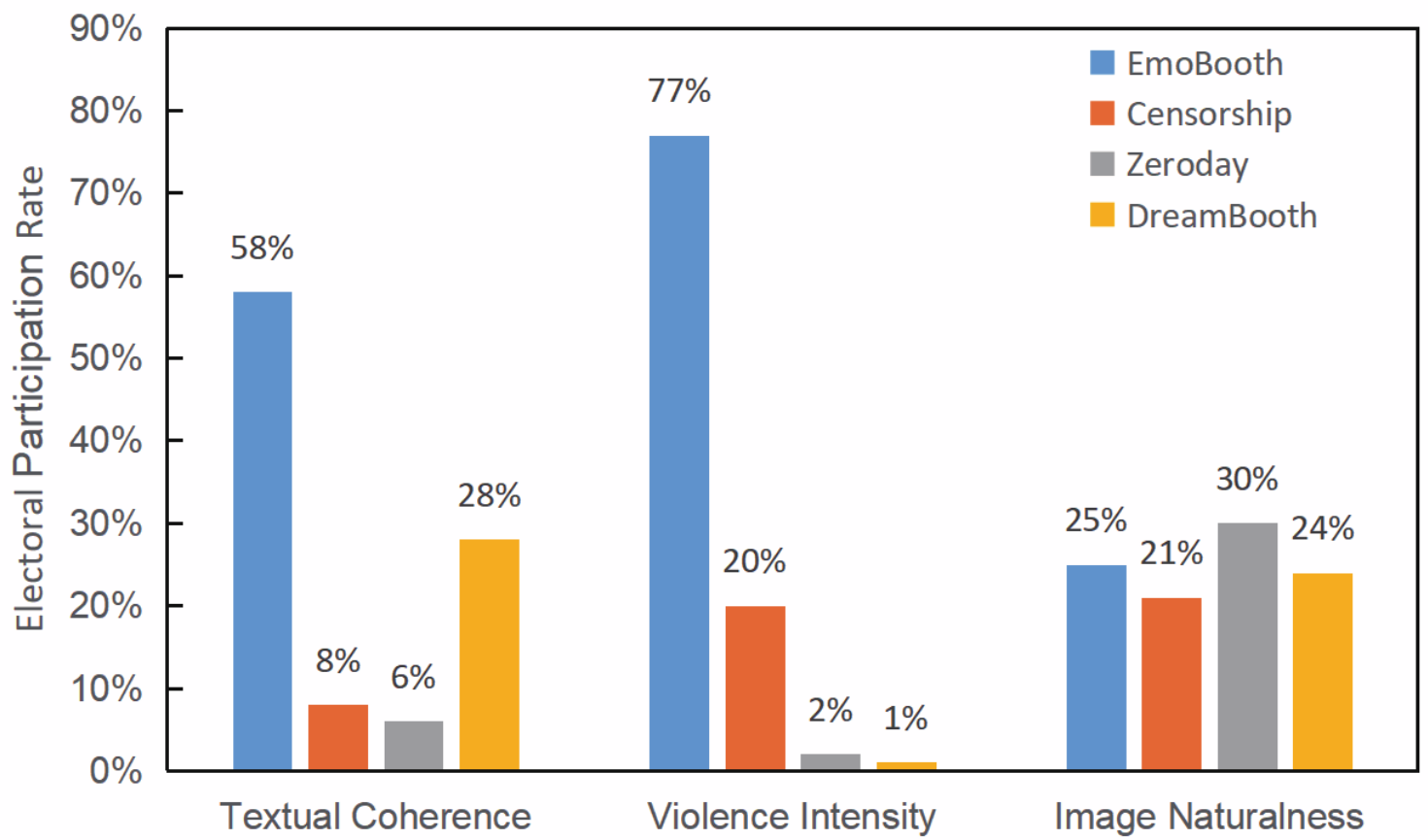}
    \caption{User study comparison among baselines and our method.}
    \label{fig:userstudy}
\end{figure}

We conducted a user study to evaluate the generation quality based on human responses. Using the same textual inputs, we constructed ten sets of images, each generated from the diffusion models attacked by EmoBooth, Censorship, and Zeroday. Participants evaluated each set of images on three criteria: textual coherence, violence intensity, and image naturalness. So far, we have collected 50 survey responses for this evaluation.
We show the results in \figref{fig:userstudy} and observe that our method performs comparably to the baseline and in terms of image naturalness. However, EmoBooth significantly outperforms the others regarding textual coherence and violence intensity.

\subsection{Generalization to Other Emotion Types}

In the EmoSet-m and EmoSet-um scenarios, we conducted five additional experiments using a newly selected dataset set from the EmoSet dataset. Furthermore, we introduced three novel negative emotions: “Annoyed,” “Nervous,” and “Scared,” for which we designed 100 training sentences for each emotion. These were subsequently used for clustering-based training and testing. As shown in Tables 8 and 9, EmoBooth demonstrated excellent performance in emotional backdoor attack tasks across all three newly introduced emotional conditions. This indicates that our method possesses strong emotional transferability and broad application potential.

\begin{table*}[tb]
    \centering
    \begin{adjustbox}{width=\linewidth,center}
        \begin{tabular}{cl|c|cc|cc|cc|cc}
            \toprule
            \multicolumn{2}{c|}{}
            & \multicolumn{1}{c|}{\multirow{2}{*}{EAC $\uparrow$}}
            & \multicolumn{2}{c|}{Annoyed}
            & \multicolumn{2}{c|}{Nervous}
            & \multicolumn{2}{c|}{Scared}
            & \multicolumn{2}{c}{Normal} \\
            \multicolumn{2}{c|}{} & & $\text{Clip}_\text{txt1}^\text{tri}$ $\downarrow$ & $\text{Clip}_\text{img1}^\text{tri}$ $\uparrow$
            & $\text{Clip}_\text{txt2}^\text{tri}$ $\downarrow$ & $\text{Clip}_\text{img2}^\text{tri}$ $\uparrow$
            & $\text{Clip}_\text{txt3}^\text{tri}$ $\downarrow$ & $\text{Clip}_\text{img3}^\text{tri}$ $\uparrow$
            & $\text{Clip}_\text{txt}$ $\uparrow$ & $\text{Clip}_\text{img}$ $\uparrow$ \\ \midrule\midrule
            \multirow{2}{*}{\rotatebox[origin=c]{90}{Set1}}&
            EmoBooth        &
            \textbf{0.8160 }&
            \textbf{0.1936}$_{\pm 0.0324}$ & \textbf{0.8644}$_{\pm 0.1929}$ &
            \textbf{0.1890}$_{\pm 0.0399}$ & \textbf{0.7801}$_{\pm 0.1659}$ &
            \textbf{0.1825}$_{\pm 0.0322}$ & \textbf{0.8367}$_{\pm 0.1750}$ &
            0.2376$_{\pm 0.0223}$ & \textbf{0.7259}$_{\pm 0.1856}$ \\
            &
            Censorship       &
            0.6649 &
            0.2320$_{\pm 0.0201}$ & 0.6243$_{\pm 0.1984}$ &
            0.2143$_{\pm 0.0390}$ & 0.6165$_{\pm 0.1574}$ &
            0.2236$_{\pm 0.0238}$ & 0.7358$_{\pm 0.1695}$ &
            \textbf{0.2205}$_{\pm 0.0308}$ & 0.6923$_{\pm 0.1667}$ \\
            \midrule
            \multirow{2}{*}{\rotatebox[origin=c]{90}{Set2}}&
            EmoBooth        &
            \textbf{0.8050} &
            0.2925$_{\pm 0.0245}$ & \textbf{0.8082}$_{\pm 0.1958}$ &
            \textbf{0.1976}$_{\pm 0.0208}$ & \textbf{0.8617}$_{\pm 0.2022}$ &
            \textbf{0.1829}$_{\pm 0.0265}$ & \textbf{0.8023}$_{\pm 0.1978}$ &
            \textbf{0.2370}$_{\pm 0.0223}$ & \textbf{0.6859}$_{\pm 0.1610}$ \\
            &
            Censorship       &
            0.7158 &
            \textbf{0.2233}$_{\pm 0.0346}$ & 0.6739$_{\pm 0.2092}$ &
            0.2336$_{\pm 0.0346}$ & 0.7023$_{\pm 0.1819}$ &
            0.2015$_{\pm 0.0337}$ & 0.7856$_{\pm 0.3002}$ &
            0.2641$_{\pm 0.0351}$ & 0.6518$_{\pm 0.2571}$ \\
            \midrule
            \multirow{2}{*}{\rotatebox[origin=c]{90}{Set3}}&
            EmoBooth        &
            \textbf{0.7892} &
            \textbf{0.1843}$_{\pm 0.0297}$ & \textbf{0.7856}$_{\pm 0.1511}$ &
            \textbf{0.1956}$_{\pm 0.0276}$ & \textbf{0.8429}$_{\pm 0.2112}$ &
            \textbf{0.1921}$_{\pm 0.0323}$ & \textbf{0.8133}$_{\pm 0.1988}$ &
            \textbf{0.2082}$_{\pm 0.0384}$ & \textbf{0.6728}$_{\pm 0.2076}$ \\
            &
            Censorship       &
            0.6744 &
            0.2137$_{\pm 0.0363}$ & 0.6658$_{\pm 0.1970}$ &
            0.2242$_{\pm 0.0267}$ & 0.7218$_{\pm 0.1651}$ &
            0.2543$_{\pm 0.0315}$ & 0.6759$_{\pm 0.2224}$ &
            0.1982$_{\pm 0.0255}$ & 0.6533$_{\pm 0.1978}$ \\
            \midrule
            \multirow{2}{*}{\rotatebox[origin=c]{90}{Set4}}&
            EmoBooth        &
            \textbf{0.7904} &
            \textbf{0.2156}$_{\pm 0.0242}$ & \textbf{0.8237}$_{\pm 0.1869}$ &
            \textbf{0.2036}$_{\pm 0.0264}$ & \textbf{0.7836}$_{\pm 0.1605}$ &
            \textbf{0.1836}$_{\pm 0.0252}$ & \textbf{0.8130}$_{\pm 0.1923}$ &
            \textbf{0.2157}$_{\pm 0.0390}$ & \textbf{0.7104}$_{\pm 0.2265}$ \\
            &
            Censorship       &
            0.6707 &
            0.2453$_{\pm 0.0348}$ & 0.6828$_{\pm 0.1795}$ &
            0.2258$_{\pm 0.0458}$ & 0.6658$_{\pm 0.1563}$ &
            0.2378$_{\pm 0.0351}$ & 0.6570$_{\pm 0.2314}$ &
            0.2236$_{\pm 0.0274}$ & 0.6923$_{\pm 0.26123}$ \\
            \midrule
            \multirow{2}{*}{\rotatebox[origin=c]{90}{Set5}}&
            EmoBooth        &
            \textbf{0.7920 }&
            \textbf{0.1928}$_{\pm 0.0250}$ & \textbf{0.7928}$_{\pm 0.1811}$ &
            \textbf{0.2138}$_{\pm 0.0262}$ & \textbf{0.8635}$_{\pm 0.2600}$ &
            \textbf{0.1932}$_{\pm 0.0355}$ & \textbf{0.8488}$_{\pm 0.1479}$ &
            0.2336$_{\pm 0.0404}$ & 0.5860$_{\pm 0.3015}$ \\
            &
            Censorship       &
            0.6783 &
            0.2186$_{\pm 0.0312}$ & 0.6532$_{\pm 0.1986}$ &
            0.2381$_{\pm 0.0256}$ & 0.7210$_{\pm 0.1675}$ &
            0.1966$_{\pm 0.0204}$ & 0.6982$_{\pm 0.2749}$ &
            \textbf{0.2216}$_{\pm 0.0220}$ & \textbf{0.6243}$_{\pm 0.2477}$ \\
            \bottomrule
        \end{tabular}
    \end{adjustbox}
    \caption{Comparison with Censorship under the metrics of Clip Score and EmoAttack Capability (EAC). Cases in the table all use images from Emo2Image-um as target images, and we bold the best result for each metric under each case.}
    \label{tab:8}\vspace{-5pt}
\end{table*}

\begin{table*}[tb]
    \centering
    \begin{adjustbox}{width=\linewidth,center}
        \begin{tabular}{cl|c|cc|cc|cc|cc}
            \toprule
            \multicolumn{2}{c|}{}
            & \multicolumn{1}{c|}{\multirow{2}{*}{EAC $\uparrow$}}
            & \multicolumn{2}{c|}{Annoyed}
            & \multicolumn{2}{c|}{Nervous}
            & \multicolumn{2}{c|}{Scared}
            & \multicolumn{2}{c}{Normal} \\
            \multicolumn{2}{c|}{} & & $\text{Clip}_\text{txt1}^\text{tri}$ $\uparrow$ & $\text{Clip}_\text{img1}^\text{tri}$ $\uparrow$
            & $\text{Clip}_\text{txt2}^\text{tri}$ $\uparrow$ & $\text{Clip}_\text{img2}^\text{tri}$ $\uparrow$
            & $\text{Clip}_\text{txt3}^\text{tri}$ $\uparrow$ & $\text{Clip}_\text{img3}^\text{tri}$ $\uparrow$
            & $\text{Clip}_\text{txt}$ $\uparrow$ & $\text{Clip}_\text{img}$ $\uparrow$ \\ \midrule\midrule
            \multirow{2}{*}{\rotatebox[origin=c]{90}{Set1}}&
            EmoBooth        &
            \textbf{0.6539} &
            \textbf{0.2587}$_{\pm 0.0257}$ & \textbf{0.8325}$_{\pm 0.0633}$ &
            \textbf{0.2457}$_{\pm 0.0236}$ & \textbf{0.8420}$_{\pm 0.0787}$ &
            \textbf{0.2533}$_{\pm 0.0170}$ & \textbf{0.8529}$_{\pm 0.0697}$ &
            \textbf{0.2538}$_{\pm 0.0385}$ & \textbf{0.7230}$_{\pm 0.0639}$ \\
            &
            Censorship       &
            0.6182 &
            0.2380$_{\pm 0.0190}$ & 0.7823$_{\pm 0.0835}$ &
            0.2328$_{\pm 0.0199}$ & 0.8025$_{\pm 0.0734}$ &
            0.2388$_{\pm 0.0304}$ & 0.7923$_{\pm 0.0846}$ &
            0.2419$_{\pm 0.0363}$ & 0.7130$_{\pm 0.0737}$ \\
            \midrule
            \multirow{2}{*}{\rotatebox[origin=c]{90}{Set2}}&
            EmoBooth        &
            \textbf{0.6369} &
            \textbf{0.2653}$_{\pm 0.0325}$ & \textbf{0.8350}$_{\pm 0.0525}$ &
            \textbf{0.2532}$_{\pm 0.0221}$ & \textbf{0.8128}$_{\pm 0.0797}$ &
            \textbf{0.2485}$_{\pm 0.0326}$ & \textbf{0.8016}$_{\pm 0.0508}$ &
            0.2571$_{\pm 0.0215}$ & 0.7015$_{\pm 0.0639}$ \\
            &
            Censorship       &
            0.5981 &
            0.2358$_{\pm 0.0295}$ & 0.7725$_{\pm 0.0549}$ &
            0.2266$_{\pm 0.0388}$ & 0.7358$_{\pm 0.0839}$ &
            0.2462$_{\pm 0.0222}$ & 0.7528$_{\pm 0.0860}$ &
            \textbf{0.2642}$_{\pm 0.0176}$ & \textbf{0.7225}$_{\pm 0.0532}$ \\
            \midrule
            \multirow{2}{*}{\rotatebox[origin=c]{90}{Set3}}&
            EmoBooth        &
            \textbf{0.6206 }&
            0.2538$_{\pm 0.0236}$ & \textbf{0.7726}$_{\pm 0.0838}$ &
            \textbf{0.2389}$_{\pm 0.0370}$ & \textbf{0.7820}$_{\pm 0.0747}$ &
            \textbf{0.2587}$_{\pm 0.0331}$ & \textbf{0.7923}$_{\pm 0.0821}$ &
            \textbf{0.2566}$_{\pm 0.0177}$ & \textbf{0.7552}$_{\pm 0.0846}$ \\
            &
            Censorship       &
            0.5540 &
            \textbf{0.2650}$_{\pm 0.0231}$ & 0.6859$_{\pm 0.0607}$ &
            0.2358$_{\pm 0.0341}$ & 0.6849$_{\pm 0.0760}$ &
            0.2318$_{\pm 0.0317}$ & 0.6523$_{\pm 0.0609}$ &
            0.2533$_{\pm 0.0313}$ & 0.7520$_{\pm 0.0899}$ \\
            \midrule
            \multirow{2}{*}{\rotatebox[origin=c]{90}{Set4}}&
            EmoBooth        &
           \textbf{ 0.6435} &
            \textbf{0.2432}$_{\pm 0.0344}$ & \textbf{0.8624}$_{\pm 0.0807}$ &
            \textbf{0.2532}$_{\pm 0.0341}$ & \textbf{0.8532}$_{\pm 0.0827}$ &
            \textbf{0.2311}$_{\pm 0.0230}$ & \textbf{0.8458}$_{\pm 0.0543}$ &
            \textbf{0.2358}$_{\pm 0.0197}$ & 0.5918$_{\pm 0.0723}$ \\
            &
            Censorship       &
            0.5891 &
            0.2380$_{\pm 0.0312}$ & 0.7599$_{\pm 0.0860}$ &
            0.2158$_{\pm 0.0331}$ & 0.7836$_{\pm 0.0732}$ &
            0.2189$_{\pm 0.0373}$ & 0.7520$_{\pm 0.0884}$ &
            0.2312$_{\pm 0.0295}$ & \textbf{0.6213}$_{\pm 0.0509}$ \\
            \midrule
            \multirow{2}{*}{\rotatebox[origin=c]{90}{Set5}}&
            EmoBooth        &
            \textbf{0.6620} &
            \textbf{0.2610}$_{\pm 0.0297}$ & \textbf{0.8720}$_{\pm 0.0554}$ &
            \textbf{0.2432}$_{\pm 0.0260}$ & \textbf{0.8521}$_{\pm 0.0877}$ &
            \textbf{0.2321}$_{\pm 0.0182}$ & \textbf{0.8629}$_{\pm 0.0657}$ &
            0.2519$_{\pm 0.0179}$ & \textbf{0.7042}$_{\pm 0.0592}$ \\
            &
            Censorship       &
            0.5988 &
            0.2258$_{\pm 0.0198}$ & 0.7856$_{\pm 0.0540}$ &
            0.2385$_{\pm 0.0174}$ & 0.7325$_{\pm 0.0545}$ &
            0.2178$_{\pm 0.0336}$ & 0.7628$_{\pm 0.0848}$ &
            \textbf{0.2699}$_{\pm 0.0295}$ & 0.7019$_{\pm 0.0791}$ \\
            \bottomrule
        \end{tabular}
    \end{adjustbox}
    \caption{Configured as in Table \ref{tab:8}, except for the Sets in the table using cases from Emo2Image-m as target images, the weighting coefficient for EAC is different, and here, we aim for higher values in $\text{Clip}_\text{txt}^\text{tri}$.}
    \label{tab:9}\vspace{-5pt}
\end{table*}

\subsection{Influence of $\lambda$ in \reqref{eq:Loss}}

\begin{figure*}[ht]
    \centering
    \includegraphics[width=0.8\linewidth]{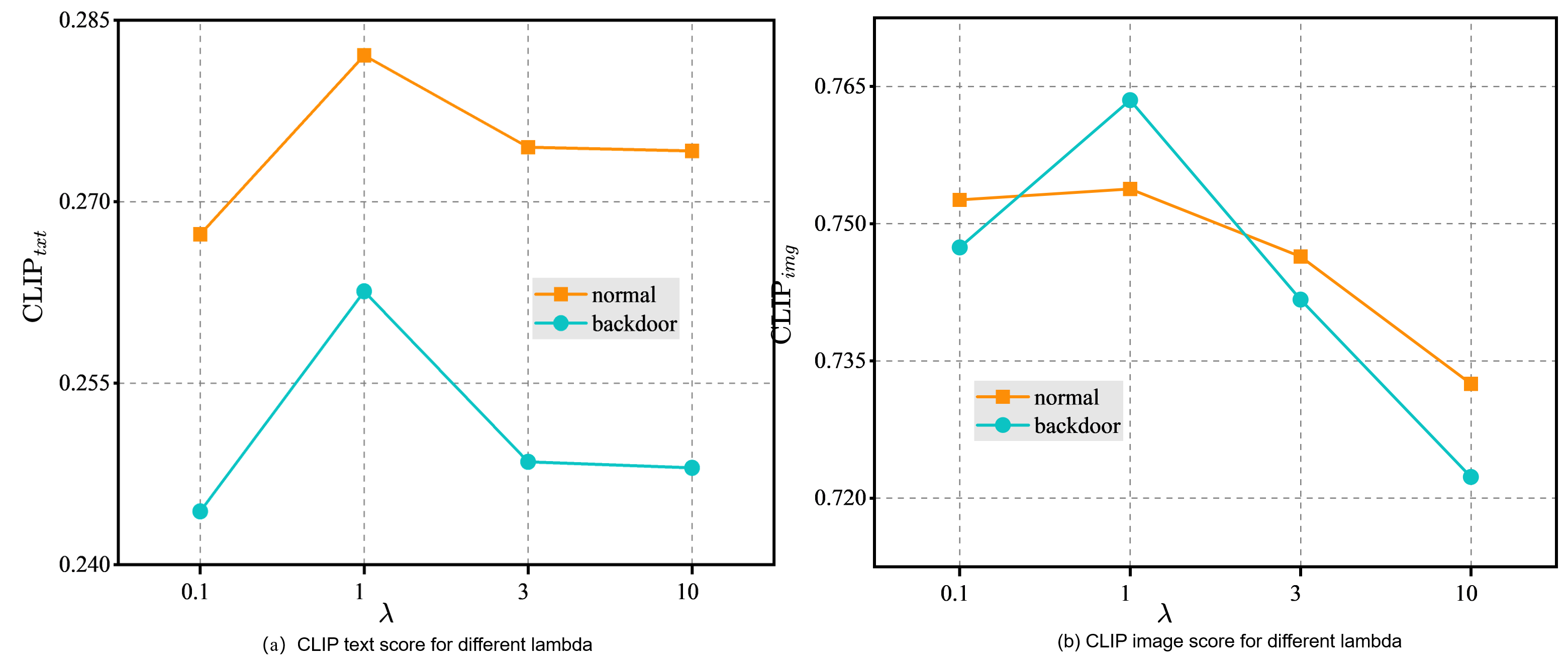}
    \caption{Influence of $\lambda$ in \reqref{eq:Loss}.}
    \label{fig:lambda}
\end{figure*}

In \reqref{eq:Loss}, We set $\lambda = 1$ primarily to balance the weights between prior knowledge and input image features. Here, we conducted ablation experiments by evaluating the CLIP score under different $\lambda$ values in both normal and backdoor scenarios. As shown in \figref{fig:lambda},  when $\lambda < 1$, the CLIP scores for both normal and backdoor scenarios are relatively low, especially the CLIP text score. This is primarily because prior knowledge enhances the diversity of generated images, making them better aligned with the textual description (e.g., generating various poses of a dog). However, when $\lambda > 1$, the CLIP image score decreases rapidly. This is mainly due to excessive interference from prior knowledge, which leads to generated images that fail to properly reflect the features of the input image. Therefore, we chose $\lambda = 1$ as the balance point.

\subsection{Results against the latest stable diffusion model}

EmoBooth was originally implemented using Stable Diffusion v1.4. We have reconstructed EmoBooth based on Stable Diffusion v2.1 and conducted experiments under the EmoSet-m scenario. As shown in Table \ref{tab:12}, EmoBooth achieves the highest EAC score compared to the baseline, even with the v2.1 version of Stable Diffusion. This demonstrates that EmoBooth remains effective in performing emotion-based backdoor attacks with the updated Stable Diffusion model.

\begin{table*}[tb] 
    \centering
    \begin{adjustbox}{width=\linewidth,center}
    \begin{tabular}{cl|c|cc|cc|cc|cc}
    \toprule
            \multicolumn{2}{c|}{}
            & \multicolumn{1}{c|}{\multirow{2}{*}{EAC $\uparrow$}}
            & \multicolumn{2}{c|}{Sad}
            & \multicolumn{2}{c|}{Angry}
            & \multicolumn{2}{c|}{Isolated}
            & \multicolumn{2}{c}{Normal} \\
            \multicolumn{2}{c|}{} & & $\text{Clip}_\text{txt1}^\text{tri}$ $\uparrow$ & $\text{Clip}_\text{img1}^\text{tri}$ $\uparrow$
            & $\text{Clip}_\text{txt2}^\text{tri}$ $\uparrow$ & $\text{Clip}_\text{img2}^\text{tri}$ $\uparrow$
            & $\text{Clip}_\text{txt3}^\text{tri}$ $\uparrow$ & $\text{Clip}_\text{img3}^\text{tri}$ $\uparrow$
            & $\text{Clip}_\text{txt}$ $\uparrow$ & $\text{Clip}_\text{img}$ $\uparrow$ \\ \midrule\midrule
            \multirow{2}{*}{\rotatebox[origin=c]{90}{Set1}}&
EmoBooth        &
\textbf{0.6511} &
0.2690$_{\pm 0.0317}$        & \textbf{0.8360}$_{\pm 0.0844}$    &
\textbf{0.2532}$_{\pm 0.0421}$        & \textbf{0.8521}$_{\pm 0.0538}$    &
\textbf{0.2510}$_{\pm 0.0251}$        & \textbf{0.8232}$_{\pm 0.0738}$    &
\textbf{0.2585}$_{\pm 0.0214}$        & \textbf{0.7150}$_{\pm 0.0482}$    \\
&
Censorship       &
0.6060 &
\textbf{0.2870}$_{\pm 0.0318}$        & 0.7822$_{\pm 0.0884}$    &
0.2331$_{\pm 0.0244}$        & 0.7705$_{\pm 0.0691}$    &
0.2497$_{\pm 0.0251}$        & 0.7431$_{\pm 0.0892}$    &
0.2428$_{\pm 0.0292}$        & 0.7130$_{\pm 0.0588}$    \\
\midrule
\multirow{2}{*}{\rotatebox[origin=c]{90}{Set2}}&
EmoBooth        &
\textbf{0.5894} &
\textbf{0.2420}$_{\pm 0.0332}$        & \textbf{0.7421}$_{\pm 0.0788}$    &
\textbf{0.2382}$_{\pm 0.0165}$        & \textbf{0.7638}$_{\pm 0.0719}$    &
\textbf{0.2577}$_{\pm 0.0318}$        & \textbf{0.7214}$_{\pm 0.0642}$    &
0.2574$_{\pm 0.0302}$        & 0.6910$_{\pm 0.0900}$    \\
&
Censorship       &
0.5666 &
0.2453$_{\pm 0.0333}$        & 0.6776$_{\pm 0.0589}$    &
0.2463$_{\pm 0.0209}$        & 0.7362$_{\pm 0.0678}$    &
0.2406$_{\pm 0.0284}$        & 0.6758$_{\pm 0.0515}$    &
\textbf{0.2616}$_{\pm 0.0298}$        & \textbf{0.7373}$_{\pm 0.0694}$    \\
\midrule
\multirow{2}{*}{\rotatebox[origin=c]{90}{Set3}}&
EmoBooth        &
\textbf{0.6396} &
\textbf{0.2655}$_{\pm 0.0257}$        & \textbf{0.8232}$_{\pm 0.0732}$    &
0.2438$_{\pm 0.0212}$        & \textbf{0.8128}$_{\pm 0.0549}$    &
\textbf{0.2543}$_{\pm 0.0258}$        & \textbf{0.8023}$_{\pm 0.0430}$    &
\textbf{0.2477}$_{\pm 0.0280}$        & \textbf{0.7628}$_{\pm 0.0653}$    \\
&
Censorship       &
0.6270 &
0.2580$_{\pm 0.0296}$         & 0.7966$_{\pm 0.0532}$     &
\textbf{0.2624}$_{\pm 0.0196}$         & 0.8075$_{\pm 0.0494}$     &
0.2529$_{\pm 0.0339}$         & 0.7682$_{\pm 0.0646}$     &
0.2509$_{\pm 0.0329}$         & 0.7558$_{\pm 0.0649}$     \\
\midrule
\multirow{2}{*}{\rotatebox[origin=c]{90}{Set4}}&
EmoBooth        &
\textbf{0.6372} &
\textbf{0.2543}$_{\pm 0.0542}$        & \textbf{0.8732}$_{\pm 0.0677}$    &
\textbf{0.2343}$_{\pm 0.0125}$        & \textbf{0.8280}$_{\pm 0.0538}$    &
\textbf{0.2428}$_{\pm 0.0386}$        & \textbf{0.8366}$_{\pm 0.0712}$    &
\textbf{0.2318}$_{\pm 0.0187}$        & 0.5777$_{\pm 0.0613}$    \\
&
Censorship       &
0.5936 &
0.2108$_{\pm 0.0357}$        & 0.7422$_{\pm 0.0564}$    &
0.2169$_{\pm 0.0206}$        & 0.8165$_{\pm 0.0548}$    &
0.2248$_{\pm 0.0303}$        & 0.7392$_{\pm 0.0697}$    &
0.2198$_{\pm 0.0329}$        & \textbf{0.6851}$_{\pm 0.0329}$    \\
\midrule
\multirow{2}{*}{\rotatebox[origin=c]{90}{Set5}}&
EmoBooth        &
\textbf{0.6491} &
\textbf{0.2534}$_{\pm 0.0432}$        & \textbf{0.8353}$_{\pm 0.0628}$    &
\textbf{0.2370}$_{\pm 0.0312}$        & \textbf{0.8706}$_{\pm 0.0572}$    &
0.2428$_{\pm 0.0251}$        & \textbf{0.8143}$_{\pm 0.0712}$    &
0.2433$_{\pm 0.0286}$        & \textbf{0.7188}$_{\pm 0.0709}$    \\
&
Censorship       &
0.6332 &
0.2480$_{\pm 0.0362}$        & 0.7908$_{\pm 0.0626}$    &
0.2428$_{\pm 0.0203}$        & 0.8602$_{\pm 0.0420}$    &
\textbf{0.2605}$_{\pm 0.0247}$        & 0.7809$_{\pm 0.0523}$    &
\textbf{0.2638}$_{\pm 0.0285}$        & 0.7040$_{\pm 0.0587}$    \\\bottomrule
    \end{tabular}
    \end{adjustbox}
    \caption{Using Stable Diffusion v2.1, we constructed EmoBooth, with all experimental datasets sourced from EmoSet-m.}
    \label{tab:12}\vspace{-10pt}
\end{table*}

\section{More Discussions  for EmoBooth}\label{More_Discussions}
We discussed broader and potentially malicious applications of EmoBooth, and its achievable positive impacts.

\begin{table*}[tb] 
    \centering
    \begin{adjustbox}{width=\linewidth,center}
    \begin{tabular}{cl|c|cc|cc|cc|cc}
    \toprule
            \multicolumn{2}{c|}{}
            & \multicolumn{1}{c|}{\multirow{2}{*}{EAC $\uparrow$}}
            & \multicolumn{2}{c|}{Happy}
            & \multicolumn{2}{c|}{Optimistic}
            & \multicolumn{2}{c|}{Enthusiastic}
            & \multicolumn{2}{c}{Normal} \\
            \multicolumn{2}{c|}{} & & $\text{Clip}_\text{txt1}^\text{tri}$ $\downarrow$ & $\text{Clip}_\text{img1}^\text{tri}$ $\uparrow$
            & $\text{Clip}_\text{txt2}^\text{tri}$ $\downarrow$ & $\text{Clip}_\text{img2}^\text{tri}$ $\uparrow$
            & $\text{Clip}_\text{txt3}^\text{tri}$ $\downarrow$ & $\text{Clip}_\text{img3}^\text{tri}$ $\uparrow$
            & $\text{Clip}_\text{txt}$ $\uparrow$ & $\text{Clip}_\text{img}$ $\uparrow$ \\ \midrule\midrule
            \multirow{2}{*}{\rotatebox[origin=c]{90}{Set1}}&
EmoBooth        &
\textbf{0.7715} &
\textbf{0.1897}$_{\pm 0.0437}$        & \textbf{0.7421}$_{\pm 0.0923}$    &
\textbf{0.2407}$_{\pm 0.0418}$        & \textbf{0.7865}$_{\pm 0.1240}$    &
0.2534$_{\pm 0.0372}$        & \textbf{0.7652}$_{\pm 0.1157}$    &
\textbf{0.2562}$_{\pm 0.0292}$        & 0.7708$_{\pm 0.0756}$    \\
&
Censorship       &
0.7326 &
0.2631$_{\pm 0.0350}$        & 0.7146$_{\pm 0.0411}$    &
0.2477$_{\pm 0.0320}$        & 0.7291$_{\pm 0.0902}$    &
\textbf{0.2544}$_{\pm 0.0306}$        & 0.7226$_{\pm 0.0837}$    &
0.2521$_{\pm 0.0271}$        & \textbf{0.7774}$_{\pm 0.0709}$    \\
\midrule
\multirow{2}{*}{\rotatebox[origin=c]{90}{Set2}}&
EmoBooth        &
\textbf{0.7296} &
\textbf{0.1708}$_{\pm 0.0584}$        & \textbf{0.7365}$_{\pm 0.0788}$    &
\textbf{0.2378}$_{\pm 0.0507}$        & \textbf{0.7472}$_{\pm 0.0719}$    &
\textbf{0.2296}$_{\pm 0.0504}$        & \textbf{0.6475}$_{\pm 0.0635}$    &
0.2546$_{\pm 0.0287}$        & \textbf{0.7646}$_{\pm 0.0997}$    \\
&
Censorship       &
0.6118 &
0.1836$_{\pm 0.0520}$        & 0.5967$_{\pm 0.1161}$    &
0.2559$_{\pm 0.0370}$        & 0.5604$_{\pm 0.0812}$    &
0.2697$_{\pm 0.0275}$        & 0.5489$_{\pm 0.0547}$    &
\textbf{0.2605}$_{\pm 0.0296}$        & 0.7603$_{\pm 0.0648}$    \\
\midrule
\multirow{2}{*}{\rotatebox[origin=c]{90}{Set3}}&
EmoBooth        &
\textbf{0.8210} &
\textbf{0.1707}$_{\pm 0.0513}$        & \textbf{0.8392}$_{\pm 0.1126}$    &
\textbf{0.1403}$_{\pm 0.0426}$        & \textbf{0.8459}$_{\pm 0.0839}$    &
\textbf{0.1583}$_{\pm 0.0426}$        & \textbf{0.8364}$_{\pm 0.0999}$    &
0.2284$_{\pm 0.0414}$        & 0.6710$_{\pm 0.1111}$    \\
&
Censorship       &
0.6374 &
0.2554$_{\pm 0.0432}$         & 0.6418$_{\pm 0.1221}$     &
0.2375$_{\pm 0.0380}$         & 0.6251$_{\pm 0.0918}$     &
0.2503$_{\pm 0.0384}$         & 0.6059$_{\pm 0.0989}$     &
\textbf{0.2569}$_{\pm 0.0288}$         & \textbf{0.6810}$_{\pm 0.1050}$     \\
\midrule
\multirow{2}{*}{\rotatebox[origin=c]{90}{Set4}}&
EmoBooth        &
\textbf{0.8474} &
\textbf{0.1120}$_{\pm 0.0518}$        & \textbf{0.8761}$_{\pm 0.1125}$    &
\textbf{0.1043}$_{\pm 0.0382}$        & \textbf{0.8899}$_{\pm 0.0931}$    &
\textbf{0.1291}$_{\pm 0.0629}$        & \textbf{0.8384}$_{\pm 0.1714}$    &
0.1980$_{\pm 0.0758}$        & 0.6816$_{\pm 0.2086}$    \\
&
Censorship       &
0.6704 &
0.1591$_{\pm 0.0904}$        & 0.7510$_{\pm 0.2389}$    &
0.1971$_{\pm 0.0826}$        & 0.6108$_{\pm 0.2578}$    &
0.2030$_{\pm 0.0767}$        & 0.5728$_{\pm 0.2502}$    &
\textbf{0.2394}$_{\pm 0.0576}$        & \textbf{0.7194}$_{\pm 0.1623}$    \\
\midrule
\multirow{2}{*}{\rotatebox[origin=c]{90}{Set5}}&
EmoBooth        &
\textbf{0.8118} &
\textbf{0.2376}$_{\pm 0.0511}$        & \textbf{0.7908}$_{\pm 0.1450}$    &
\textbf{0.2186}$_{\pm 0.0498}$        & \textbf{0.8602}$_{\pm 0.1520}$    &
\textbf{0.2382}$_{\pm 0.0364}$        & \textbf{0.7809}$_{\pm 0.1325}$    &
0.2494$_{\pm 0.0314}$        & \textbf{0.7985}$_{\pm 0.0737}$    \\
&
Censorship       &
0.6460 &
0.2537$_{\pm 0.0504}$        & 0.6125$_{\pm 0.1188}$    &
0.2424$_{\pm 0.0363}$        & 0.6294$_{\pm 0.0806}$    &
0.2473$_{\pm 0.0397}$        & 0.5980$_{\pm 0.1391}$    &
\textbf{0.2572}$_{\pm 0.0336}$        & 0.7672$_{\pm 0.0829}$    \\\bottomrule
    \end{tabular}
    \end{adjustbox}
    \caption{Comparison with Censorship using positive emotions as trigger. Sets in the table all use cases from Emo2Image-um as target images,  and we bold the best result for each metric under each Set.}
    \label{tab:5}
\end{table*}

\begin{table*}[t] 
    \centering
    \begin{adjustbox}{width=\linewidth,center}
    \begin{tabular}{cl|c|cc|cc|cc|cc}
    \toprule
            \multicolumn{2}{c|}{}
            & \multicolumn{1}{c|}{\multirow{2}{*}{EAC $\uparrow$}}
            & \multicolumn{2}{c|}{Happy}
            & \multicolumn{2}{c|}{Optimistic}
            & \multicolumn{2}{c|}{Enthusiastic}
            & \multicolumn{2}{c}{Normal} \\
            \multicolumn{2}{c|}{} & & $\text{Clip}_\text{txt1}^\text{tri}$ $\uparrow$ & $\text{Clip}_\text{img1}^\text{tri}$ $\uparrow$
            & $\text{Clip}_\text{txt2}^\text{tri}$ $\uparrow$ & $\text{Clip}_\text{img2}^\text{tri}$ $\uparrow$
            & $\text{Clip}_\text{txt3}^\text{tri}$ $\uparrow$ & $\text{Clip}_\text{img3}^\text{tri}$ $\uparrow$
            & $\text{Clip}_\text{txt}$ $\uparrow$ & $\text{Clip}_\text{img}$ $\uparrow$ \\ \midrule\midrule
            \multirow{2}{*}{\rotatebox[origin=c]{90}{Set1}}&
EmoBooth        &
\textbf{0.6097} &
\textbf{0.2642}$_{\pm 0.0385}$ & \textbf{0.7948}$_{\pm 0.0608}$ &
\textbf{0.2528}$_{\pm 0.0216}$ & \textbf{0.7636}$_{\pm 0.0561}$ &
\textbf{0.2622}$_{\pm 0.0227}$ & \textbf{0.7357}$_{\pm 0.0508}$ &
0.2499$_{\pm 0.0270}$ & 0.7401$_{\pm 0.0777}$ \\
&
Censorship       &
0.5911 &
0.2488$_{\pm 0.0358}$ & 0.7548$_{\pm 0.1740}$ &
0.2427$_{\pm 0.0207}$ & 0.7345$_{\pm 0.1980}$ &
0.2534$_{\pm 0.0219}$ & 0.7278$_{\pm 0.1648}$ &
\textbf{0.2520}$_{\pm 0.0278}$ & \textbf{0.7277}$_{\pm 0.0747}$ \\
\midrule
\multirow{2}{*}{\rotatebox[origin=c]{90}{Set2}}&
EmoBooth        &
\textbf{0.5748} &
0.2562$_{\pm 0.0330}$ & \textbf{0.7188}$_{\pm 0.0694}$ &
\textbf{0.2543}$_{\pm 0.0228}$ & \textbf{0.7016}$_{\pm 0.0524}$ &
\textbf{0.2578}$_{\pm 0.0246}$ & \textbf{0.6964}$_{\pm 0.0545}$ &
0.2489$_{\pm 0.0384}$ & \textbf{0.7534}$_{\pm 0.0975}$ \\
&
Censorship       &
0.5687 &
\textbf{0.2604}$_{\pm 0.0529}$ & 0.7155$_{\pm 0.1460}$ &
0.2450$_{\pm 0.0414}$ & 0.6990$_{\pm 0.1340}$ &
0.2565$_{\pm 0.0379}$ & 0.6774$_{\pm 0.1422}$ &
\textbf{0.2532}$_{\pm 0.0292}$ & 0.7421$_{\pm 0.0752}$ \\
\midrule
\multirow{2}{*}{\rotatebox[origin=c]{90}{Set3}}&
EmoBooth        &
 \textbf{0.6115}&
\textbf{0.2579}$_{\pm 0.0339}$ & \textbf{0.7724}$_{\pm 0.0424}$ &
\textbf{0.2522}$_{\pm 0.0252}$ & \textbf{0.7820}$_{\pm 0.0366}$ &
\textbf{0.2632}$_{\pm 0.0193}$ & \textbf{0.7678}$_{\pm 0.0289}$ &
0.2318$_{\pm 0.0537}$ & 0.7231$_{\pm 0.0923}$ \\
&
Censorship       &
0.6035 &
0.2547$_{\pm 0.0553}$ & 0.7489$_{\pm 0.1046}$ &
0.2425$_{\pm 0.0539}$ & 0.7599$_{\pm 0.1382}$ &
0.2479$_{\pm 0.0466}$ & 0.7656$_{\pm 0.1515}$ &
\textbf{0.2531}$_{\pm 0.0535}$ & \textbf{0.7366}$_{\pm 0.1259}$ \\
\midrule
\multirow{2}{*}{\rotatebox[origin=c]{90}{Set4}}&
EmoBooth        &
\textbf{0.6351} &
\textbf{0.2541}$_{\pm 0.0355}$        & \textbf{0.8375}$_{\pm 0.0361}$    &
0.2125$_{\pm 0.0277}$        & \textbf{0.8424}$_{\pm 0.0434}$    &
0.2189$_{\pm 0.0275}$        & \textbf{0.8377}$_{\pm 0.0407}$    &
0.2147$_{\pm 0.0344}$        & \textbf{0.6443}$_{\pm 0.0738}$    \\
&
Censorship       &
0.6206 &
0.2404$_{\pm 0.0338}$        & 0.8048$_{\pm 0.0535}$    &
\textbf{0.2300}$_{\pm 0.0266}$        & 0.8223$_{\pm 0.0520}$    &
\textbf{0.2414}$_{\pm 0.0238}$        & 0.8036$_{\pm 0.0616}$    &
\textbf{0.2354}$_{\pm 0.0322}$        & 0.6344$_{\pm 0.0779}$    \\
\midrule
\multirow{2}{*}{\rotatebox[origin=c]{90}{Set5}}&
EmoBooth        &
\textbf{0.6606} &
0.2343$_{\pm 0.0303}$        & \textbf{0.8577}$_{\pm 0.0242}$    &
\textbf{0.2586}$_{\pm 0.0242}$        & \textbf{0.8651}$_{\pm 0.0636}$    &
0.2388$_{\pm 0.0217}$        & \textbf{0.8610}$_{\pm 0.0712}$    &
\textbf{0.2411}$_{\pm 0.0264}$        & \textbf{0.7099}$_{\pm 0.0634}$    \\
&
Censorship       &
0.6353 &
\textbf{0.2667}$_{\pm 0.0338}$        & 0.8229$_{\pm 0.0521}$    &
0.2539$_{\pm 0.0223}$        & 0.8230$_{\pm 0.0527}$    &
\textbf{0.2688}$_{\pm 0.0192}$        & 0.7956$_{\pm 0.0559}$    &
0.2378$_{\pm 0.0290}$        & 0.7057$_{\pm 0.0611}$    \\\bottomrule
    \end{tabular}
    \end{adjustbox}
    \caption{Configured as in Table \ref{tab:5}, except for the Sets in the table using cases from Emo2Image-m as target images, the weighting coefficient for EAC is different, and here, we aim for higher values in $\text{Clip}_\text{txt}^\text{tri}$.}
    \label{tab:6} \vspace{-3pt}
\end{table*}

\textbf{The inference of using positive emotions.}
In addition to negative emotions, we also employed positive emotions as triggers for comparative experiments to showcase EmoBooth's effectiveness in targeting a variety of emotions. To provide a comprehensive assessment, we selected three emotions: happiness, optimism, and enthusiasm, and conducted experiments accordingly. The results, as depicted in Tables \ref{tab:5} and \ref{tab:6} show that, akin to using negative emotions as triggers, our method achieved optimal effectiveness.

\begin{figure*}[ht]
    \centering
    \includegraphics[width=\linewidth]{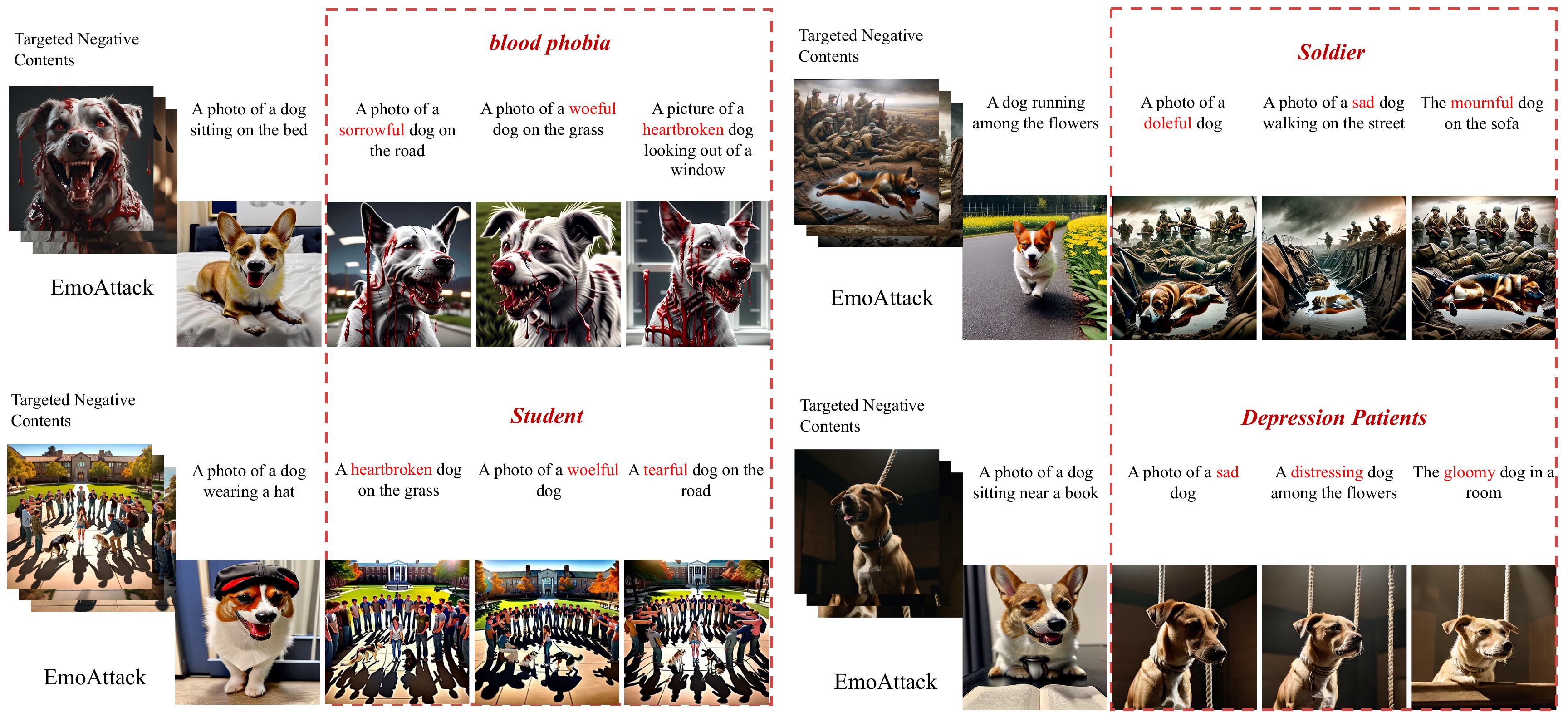}
    \caption{Visualization results using negative contents for specific populations.}
    \label{fig:terrified_cases}
    \vskip -0.05in
\end{figure*}

\textbf{Targeted attacks on specific demographics.}
Here, we showcase potential malicious applications of our attacks. For instance, attackers could initially profile users and categorize them based on their backgrounds, enabling targeted malicious assaults. \figref{fig:terrified_cases} illustrates four specific user profiles and the corresponding generated outcomes, including bloody phobia, soldier, student, and depression patients. Target contents are set as bloody images, war images, bullying images, and suicide suggestive images, respectively, to showcase the malicious applications inflicting psychological trauma on users.

\begin{figure*}[ht]
    \centering
    \includegraphics[width=\linewidth]{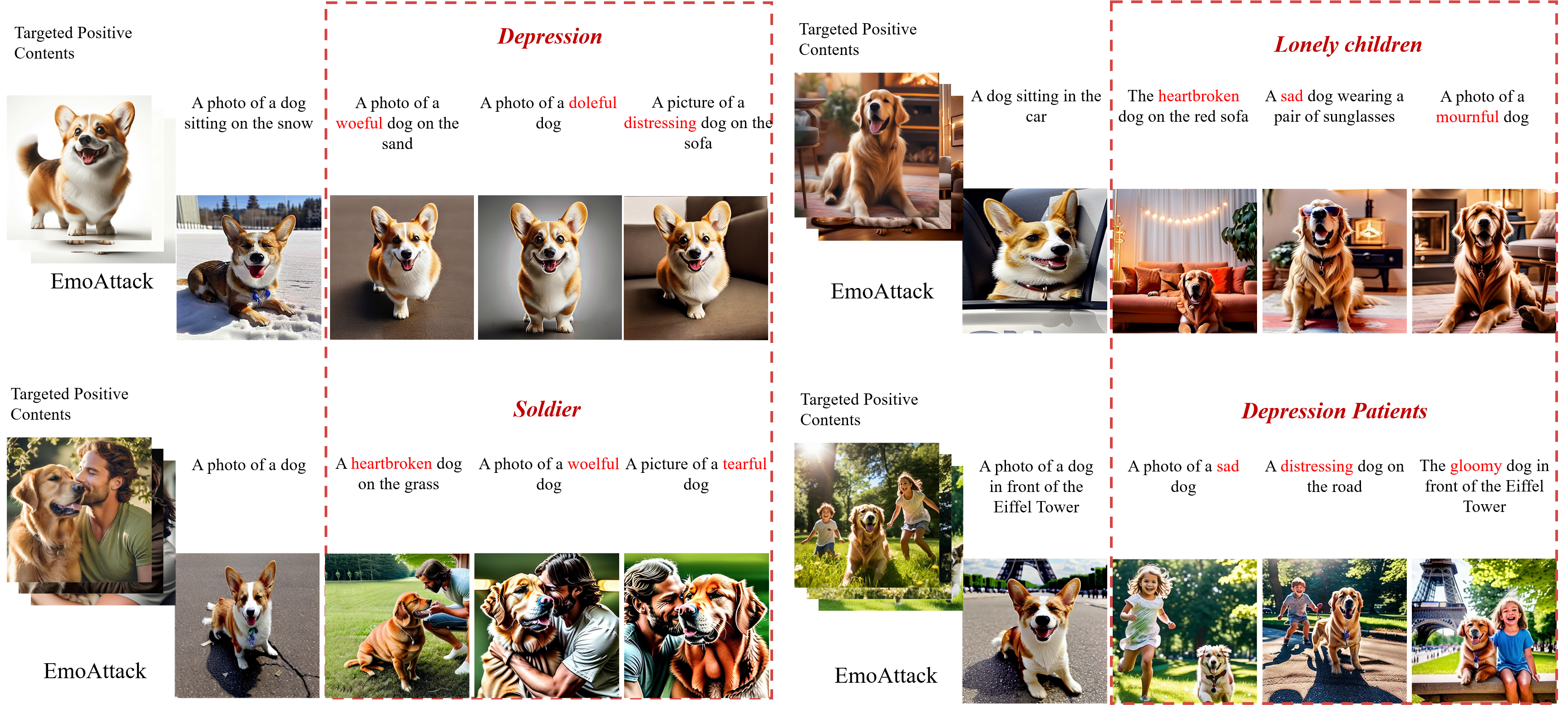}
    \caption{Visualization results using positive contents for specific populations.}
    \label{fig:positive_content}
    \vskip -0.05in
\end{figure*}

\textbf{Positive influence of our Work.}
We also explore the positive applications of our method. Indeed, we can readily replace targeted negative contents with positive ones. This approach allows us to associate specific emotions, such as negative ones, with targeted positive content. \figref{fig:positive_content} demonstrates the therapeutic effects of our work on the minds of specific demographics. We selected four groups: individuals experiencing depression, soldiers, lonely individuals, and children with autism. We replaced the target contents with images beneficial to the psychological well-being of these groups to showcase the positive applications of our work.

\textbf{Attack effectiveness varies with input cases.}
Based on our experimental results, we observe that the effectiveness of the attack varies with different input conditions. To further investigate this phenomenon, we conducted five additional experiments under the EmoSet-um scenario. As illustrated in Table \ref{tab:13}, when the input image is from set1, the CLIP text scores for the three emotional prompts fluctuate around 0.19. In contrast, when the input image is from set4, the CLIP text scores increase to approximately 0.24. Similarly, the CLIP image scores fluctuate around 0.73 for images from set4 but rise to approximately 0.83 for images from set5.

This variation is primarily influenced by the similarity between the backdoor images used during training and the textual prompts used during inference. Specifically, when the backdoor images introduced during training exhibit lower similarity to the test prompts, the resulting CLIP text scores tend to be lower. Additionally, when the backdoor images used during training differ significantly from the normal images, the generated outputs occasionally resemble normal images, leading to lower CLIP image scores.

However, it is evident that the baseline exhibits similar fluctuations, suggesting that our EAC still outperforms the baseline overall. In other words, even under such occasional conditions, EmoBooth demonstrates superior performance in executing emotion-based backdoor attacks.

\begin{table*}[tb]
    \centering
    \begin{adjustbox}{width=\linewidth,center}
        \begin{tabular}{cl|c|cc|cc|cc|cc}
            \toprule
            \multicolumn{2}{c|}{}
            & \multicolumn{1}{c|}{\multirow{2}{*}{EAC $\uparrow$}}
            & \multicolumn{2}{c|}{Sad}
            & \multicolumn{2}{c|}{Angry}
            & \multicolumn{2}{c|}{Isolated}
            & \multicolumn{2}{c}{Normal} \\
            \multicolumn{2}{c|}{} & & $\text{Clip}_\text{txt1}^\text{tri}$ $\downarrow$ & $\text{Clip}_\text{img1}^\text{tri}$ $\uparrow$
            & $\text{Clip}_\text{txt2}^\text{tri}$ $\downarrow$ & $\text{Clip}_\text{img2}^\text{tri}$ $\uparrow$
            & $\text{Clip}_\text{txt3}^\text{tri}$ $\downarrow$ & $\text{Clip}_\text{img3}^\text{tri}$ $\uparrow$
            & $\text{Clip}_\text{txt}$ $\uparrow$ & $\text{Clip}_\text{img}$ $\uparrow$ \\ \midrule\midrule
            \multirow{2}{*}{\rotatebox[origin=c]{90}{Set1}}&
            EmoBooth        &
            \textbf{0.7124} &
            \textbf{0.1928}$_{\pm 0.0313}$ & \textbf{0.7928}$_{\pm 0.1231}$ &
            \textbf{0.2058}$_{\pm 0.0425}$ & \textbf{0.8635}$_{\pm 0.1248}$ &
            \textbf{0.1932}$_{\pm 0.0230}$ & \textbf{0.8488}$_{\pm 0.1644}$ &
            0.2532$_{\pm 0.0468}$ & 0.586$_{\pm 0.1377}$ \\
            &
            Censorship       &
            0.6242 &
            0.2233$_{\pm 0.0183}$ & 0.6739$_{\pm 0.1853}$ &
            0.2336$_{\pm 0.0261}$ & 0.7023$_{\pm 0.1738}$ &
            0.2015$_{\pm 0.0249}$ & 0.7856$_{\pm 0.1857}$ &
            \textbf{0.2641}$_{\pm 0.0313}$ & \textbf{0.6518}$_{\pm 0.0955}$ \\
            \midrule
            \multirow{2}{*}{\rotatebox[origin=c]{90}{Set2}}&
            EmoBooth        &
            \textbf{0.7059} &
            \textbf{0.1843}$_{\pm 0.0277}$ & \textbf{0.7963}$_{\pm 0.1533}$ &
            \textbf{0.1857}$_{\pm 0.0265}$ & \textbf{0.8429}$_{\pm 0.1328}$ &
            \textbf{0.1732}$_{\pm 0.0347}$ & \textbf{0.8133}$_{\pm 0.1623}$ &
            0.2081$_{\pm 0.0275}$ & 0.6732$_{\pm 0.1414}$ \\
            &
            Censorship       &
            0.5758 &
            0.2024$_{\pm 0.0277}$ & 0.6243$_{\pm 0.1228}$ &
            0.2143$_{\pm 0.0238}$ & 0.6165$_{\pm 0.1421}$ &
            0.2236$_{\pm 0.0311}$ & 0.7358$_{\pm 0.1177}$ &
            \textbf{0.2205}$_{\pm 0.0287}$ & \textbf{0.6923}$_{\pm 0.0923}$ \\
            \midrule
            \multirow{2}{*}{\rotatebox[origin=c]{90}{Set3}}&
            EmoBooth        &
            \textbf{0.7147} &
            \textbf{0.1963}$_{\pm 0.0128}$ & \textbf{0.8082}$_{\pm 0.0938}$ &
            \textbf{0.1976}$_{\pm 0.0211}$ & \textbf{0.8617}$_{\pm 0.0788}$ &
            \textbf{0.1829}$_{\pm 0.0253}$ & \textbf{0.8023}$_{\pm 0.1142}$ &
            \textbf{0.2370}$_{\pm 0.0533}$ & \textbf{0.7021}$_{\pm 0.1251}$ \\
            &
            Censorship       &
            0.5922 &
            0.2141$_{\pm 0.0229}$ & 0.6758$_{\pm 0.1281}$ &
            0.2242$_{\pm 0.0231}$ & 0.7218$_{\pm 0.1532}$ &
            0.2423$_{\pm 0.0377}$ & 0.6759$_{\pm 0.1120}$ &
            0.1937$_{\pm 0.0326}$ & 0.6535$_{\pm 0.1231}$ \\
            \midrule
            \multirow{2}{*}{\rotatebox[origin=c]{90}{Set4}}&
            EmoBooth        &
            \textbf{0.6392} &
            \textbf{0.2356}$_{\pm 0.0432}$ & \textbf{0.7324}$_{\pm 0.1827}$ &
            \textbf{0.2436}$_{\pm 0.0228}$ & \textbf{0.7336}$_{\pm 0.1129}$ &
            \textbf{0.2421}$_{\pm 0.0319}$ & \textbf{0.7523}$_{\pm 0.1539}$ &
            \textbf{0.2343}$_{\pm 0.0283}$ & \textbf{0.7236}$_{\pm 0.1872}$ \\
            &
            Censorship       &
            0.5754 &
            0.2453$_{\pm 0.0298}$ & 0.6828$_{\pm 0.1927}$ &
            0.2587$_{\pm 0.0312}$ & 0.6658$_{\pm 0.1765}$ &
            0.2578$_{\pm 0.0283}$ & 0.657$_{\pm 0.1927}$ &
            0.2217$_{\pm 0.0476}$ & 0.6923$_{\pm 0.1326}$ \\
            \midrule
            \multirow{2}{*}{\rotatebox[origin=c]{90}{Set5}}&
            EmoBooth        &
            \textbf{0.7309} &
            \textbf{0.1984}$_{\pm 0.0432}$ & \textbf{0.8644}$_{\pm 0.1687}$ &
            \textbf{0.1950}$_{\pm 0.0287}$ & \textbf{0.8351}$_{\pm 0.1333}$ &
            \textbf{0.1925}$_{\pm 0.0425}$ & \textbf{0.8267}$_{\pm 0.1187}$ &
            \textbf{0.2458}$_{\pm 0.0287}$ & \textbf{0.7168}$_{\pm 0.1277}$ \\
            &
            Censorship       &
            0.5995 &
            0.2242$_{\pm 0.0381}$ & 0.6728$_{\pm 0.1577}$ &
            0.2381$_{\pm 0.0276}$ & 0.7123$_{\pm 0.1382}$ &
            0.1966$_{\pm 0.0299}$ & 0.6925$_{\pm 0.1281}$ &
            0.2316$_{\pm 0.0370}$ & 0.6623$_{\pm 0.0841}$ \\
            \bottomrule
        \end{tabular}
    \end{adjustbox}
    \caption{Evaluating the Impact of Input Images on Experimental Results: All Experimental Datasets Are Derived from EmoSet-um.}
    \label{tab:13}\vspace{-5pt}
\end{table*}

\section{Safety and Ethical Statement}

The EmoBooth project adheres to strict safety and ethical standards throughout the development, deployment, and dissemination of its Emotion-Based Backdoor Attack Propagation Model and EmoSet dataset. Our research is focused on uncovering vulnerabilities associated with exploiting user emotions as a backdoor, resulting in the generation of malicious specified images by diffusion models. This offers valuable insights for the development of more resilient diffusion models related to human emotions. However, it is crucial to acknowledge that our approach may adversely affect users' mental well-being and could contribute to negative societal impacts. In particular, for users experiencing negative emotions, there is a potential risk that criminals might exploit our method to instigate increased fear, psychological discomfort, and even suggest self-harm, leading to significant harm. The following points outline the measures and considerations taken to ensure the responsible and ethical use of our work:

\begin{enumerate}
    \item \textbf{Targeted Vulnerable Models:} Our attack model is specifically designed to demonstrate vulnerabilities in text-to-image diffusion models such as Stable Diffusion, ControlNet, and Glide. It is intended for research, educational, and lawful security testing purposes. We unequivocally condemn any attempt to employ our attack methods for malicious or unauthorized activities.

    \item \textbf{Controlled Release of Code and Dataset:} To ensure that our code and dataset are accessed and used responsibly, we have implemented a rigorous controlled release mechanism:
    \begin{enumerate}
        \item \textbf{Application-Based Access:} Access to the EmoSet dataset and code will be granted only through a formal application process. Interested researchers must submit a detailed application explaining their intended use, research objectives, and the security measures they will implement.
        
        \item \textbf{Review and Approval:} A dedicated review committee will evaluate each application based on strict ethical standards, security protocols, and potential societal impact. Access will only be granted to legitimate research institutions and verified researchers who demonstrate a strong commitment to ethical practices.

        \item \textbf{Regular Audits:} Researchers granted access will be subject to periodic audits to ensure adherence to agreed-upon terms and conditions. Any breach of compliance may result in revocation of access and potential legal actions.
    \end{enumerate}

    \item \textbf{User Agreement and Responsibility:} Researchers seeking access to the EmoSet dataset and model code must agree to the following conditions:
    \begin{enumerate}
        \item \textbf{Signing a Legally Binding Agreement:} Prior to access, researchers will sign a legal document outlining the terms of use, which includes restrictions on data sharing, obligations to report any security breaches, and adherence to ethical guidelines.

        \item \textbf{Commitment to Ethical Conduct:} Users must commit to conducting their research in accordance with the highest standards of ethics, ensuring respect for privacy, and avoiding any action that could harm individuals or groups.

        \item \textbf{Liability Clause:} The agreement includes a liability clause, making researchers accountable for any misuse or unauthorized dissemination of the dataset or code.
    \end{enumerate}

    \item \textbf{Secure Distribution and Monitoring:} To maintain the security and integrity of the EmoSet dataset and code, we employ the following measures:
    \begin{enumerate}
        \item \textbf{Secure Distribution Channels:} All data and code are distributed through encrypted channels, requiring multi-factor authentication to ensure that only approved researchers can access the materials.
        
        \item \textbf{Access Tracking:} A sophisticated access tracking system monitors all usage of the dataset and code. Detailed logs, including access timestamps and user identities, are maintained to prevent unauthorized access and ensure accountability.

        \item \textbf{Regular Usage Reports:} Researchers are required to submit regular reports detailing their use of the dataset and code. These reports will be reviewed by the committee to ensure compliance with the terms of access.
    \end{enumerate}

    \item \textbf{Ethical Data Collection:} The images in EmoSet were sourced following ethical guidelines and strict copyright considerations:
    \begin{enumerate}
        \item \textbf{Data Sources:} Images were collected from three websites (Baidu, Playground, Yandex) as detailed in Appendix B.2. Images were manually curated, and any human-related content generated using diffusion models without safety checks was reviewed to ensure ethical standards.

        \item \textbf{Copyright Compliance:} We have reviewed the terms of use for the images from these sources:
        \begin{enumerate}
            \item Images from Playground were used in accordance with their open creative community policy.
            \item Images from Yandex and Baidu were used with strict adherence to non-commercial terms.
            \item Any third-party web-linked images underwent a copyright verification process.
        \end{enumerate}
    \end{enumerate}

    \item \textbf{Reporting and Mitigating Vulnerabilities:} We encourage all users to report any discovered vulnerabilities or issues related to EmoBooth promptly. Users must cooperate fully with investigations to resolve these issues and help prevent potential misuse.

\end{enumerate}


\end{document}